\documentclass[10pt,twocolumn,letterpaper]{article}

\usepackage[pagenumbers,evaluations]{wacv}

\usepackage{graphicx}
\usepackage{xcolor}
\usepackage{colortbl}
\usepackage{tcolorbox}
\newtcolorbox{promptbox}{%
  colback=black!5, colframe=black!30,
  boxrule=0.4pt, arc=1.5pt,
  left=5pt, right=5pt, top=4pt, bottom=4pt,
  before skip=5pt, after skip=5pt,
  fontupper=\small}
\usepackage{amssymb}
\usepackage{amsmath}
\usepackage{booktabs}
\usepackage{multirow}
\usepackage{float}
\usepackage[shortlabels]{enumitem}
\usepackage{siunitx}
\usepackage{comment}
\usepackage{stfloats}

\newcommand{\imnet}[1][1]{ImageNet-#1k}
\newcommand{\nood}{\nbench\xspace}

\definecolor{better_gray}{HTML}{72777D}
\newcommand{\g}[1]{{\scriptsize\color{better_gray}#1}}

\definecolor{predcol}{RGB}{40,90,200}
\definecolor{oodcol}{RGB}{200,70,20}

\newcommand{\bench}{\textsc{\mbox{Fi-ImageNet}-1k}\xspace}

\newcommand{\nflagged}{2597}
\newcommand{\nnosure}{2123}
\newcommand{\nempty}{474}

\newcommand{\npoolid}{962}
\newcommand{\nambig}{980}

\newcommand{\nbench}{655}

\newcommand{\nbenchlab}{522}
\newcommand{\nid}{48{,}365}

\definecolor{wacvblue}{rgb}{0.21,0.49,0.74}
\usepackage[pagebackref,breaklinks,colorlinks,allcolors=wacvblue]{hyperref}

\def\wacvPaperID{686}
\def\confName{WACV}
\def\confYear{2027}

\title{Fi-ImageNet-1k:\\ 
An OOD benchmark From the Inside of the ImageNet-1k validation set}

\author{
    Ruslan Rozumnyi, Mat\v{e}j Such\'{a}nek, Tom\'{a}\v{s} Voj\'{i}\v{r}, Kl\'{a}ra Janou\v{s}kov\'{a}, Ji\v{r}\'{i} Matas
    \vspace{2mm}\\
    Visual Recognition Group\\
    Department of Cybernetics\\
    Faculty of Electrical Engineering\\
    Czech Technical University in Prague\\
}

\begin{document}
\maketitle

\begin{abstract}

Out-of-distribution (OOD) detection predicts whether a test image belongs to none of the predefined classes. 
To evaluate this task, benchmarks need images from outside the in-distribution (ID) data; typically, these are defined or collected in an ad hoc fashion.
Since no ground truth is perfect, ID-labeled datasets themselves contain a natural source of OOD images. 
We exploit such annotation errors and present \bench{}, an OOD dataset built from ImageNet-1k validation images that the recent ReImageNet reannotation effort assigned to no ImageNet-1k class.
Each image was examined by expert human annotators supported by evidence from MLLMs, VLMs, and reverse image search, comparing it against all visually similar ID classes.
We keep only images that could be assigned a specific class \emph{outside} the ImageNet-1k label space.

The resulting \bench{}, with \nbench{}~images from \nbenchlab{}~classes, is substantially more challenging than any commonly used OOD dataset.
No evaluated combination of classifier and OOD detector achieves a false positive rate below 51\% at 95\% true positive rate (FPR@95).
Compared to the recent NINCO, our dataset is 3.8\texttimes{} more challenging in the FPR@95 metric for state-of-the-art supervised OOD detection methods. The dataset will be released upon acceptance.

\end{abstract}

\section{Introduction}
\label{sec:intro}

Open-set detection entails the task of identifying, at test time, images that belong to none of a predefined set of training classes\footnote{Unless zero-shot detection is considered.}.
The term out-of-distribution (OOD) detection is often used interchangeably, especially when the OOD data share visual similarities to images from the predefined classes.
OOD detection is a prerequisite for deploying models safely in the real world, where inputs cannot be guaranteed to originate from a fixed set of classes \cite{hendrycks2017msp}.

Yet, OOD detection has not been an integral component of 
state-of-the-art models for classification tasks, 
such as EVA-02 \cite{fang2024eva02} and \mbox{ConvNeXt V2} \cite{liu2022convnext}.
This also holds for detection methods based on zero-shot vision–language models (VLMs), such as CLIP \cite{radford2021clip} and SigLIP~2 \cite{tschannen2025siglip2}, which are deployed without a mechanism for recognizing inputs that fall outside their respective label space
\cite{ming2022mcm, wang2023clipn}.

\begin{figure}[t]
  \centering
  \framebox{
  \includegraphics[width=0.92\columnwidth]{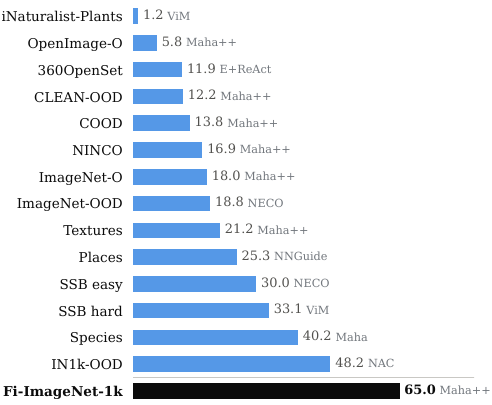}
  }
  \caption{
  Results for the best-performing EVA02-L model on popular OOD benchmarks that use ImageNet-1k as the in-distribution data, using FPR@95 ($\downarrow$), in \%.
  Bars show the \emph{best} (right, in gray) of the 21 post-hoc OOD detectors on a given benchmark.
  Even for the per-dataset top choice, the FPR@95 on the \bench{} benchmark is 1.3\texttimes{} higher than on the hardest of the other benchmarks (IN1k-OOD) and 3.8\texttimes{} higher than on recent NINCO.
  }
  \label{fig:fpr_ours_vs_popular}
\end{figure}
\newcommand{\teasersubfigure}[3]{%
\begin{minipage}[t]{0.137\linewidth}\centering
  \includegraphics[width=\linewidth]{figs/teaser/#1.jpg}\\[-3pt]
  {\small\textcolor{green!50!black}{#2}}\\[-2pt]
  {\small\textcolor{red}{#3}}
\end{minipage}%
}
\begin{figure*}[t]
  \begin{minipage}{\textwidth}
  \teasersubfigure{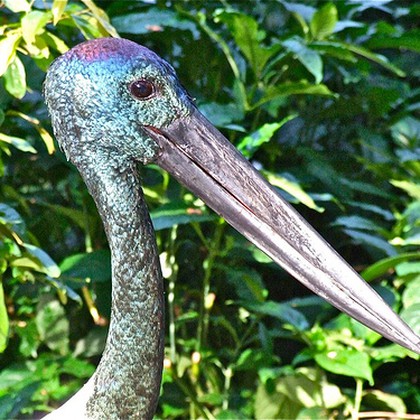}{black-necked stork}{black stork}
  \teasersubfigure{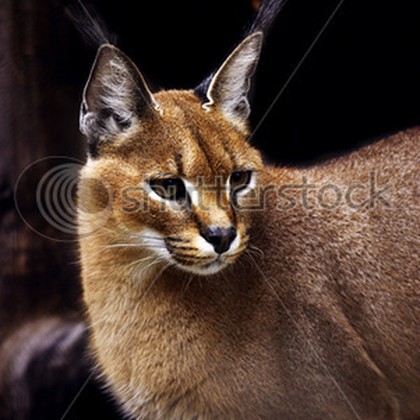}{caracal}{lynx}
  \teasersubfigure{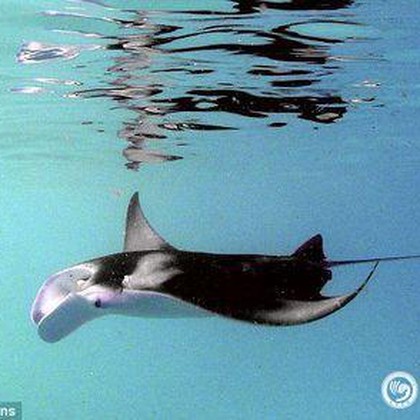}{manta ray}{electric ray}
  \teasersubfigure{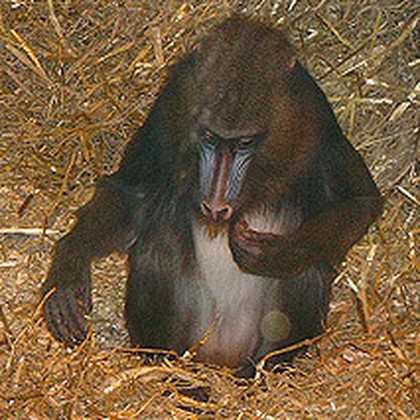}{mandrill}{siamang}
  \teasersubfigure{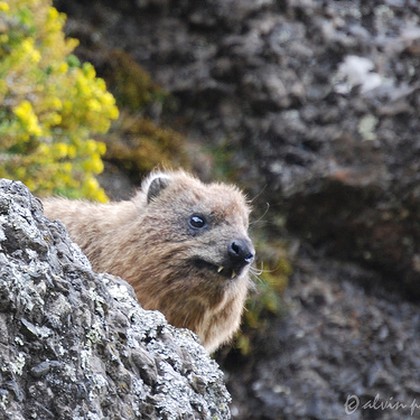}{hyrax}{marmot}
  \teasersubfigure{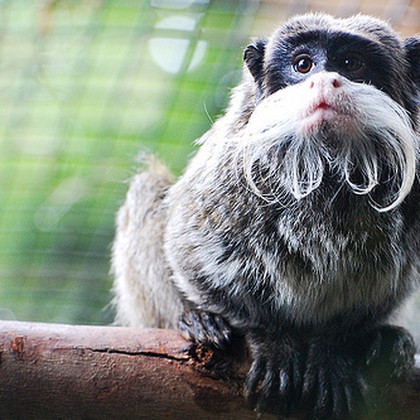}{emperor tamarin}{titi monkey}
  \teasersubfigure{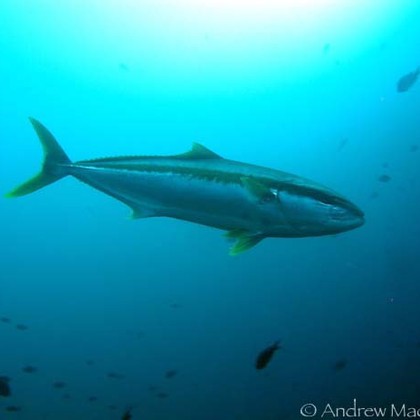}{greater amberjack}{northern pike}
  \\[2ex]
  \teasersubfigure{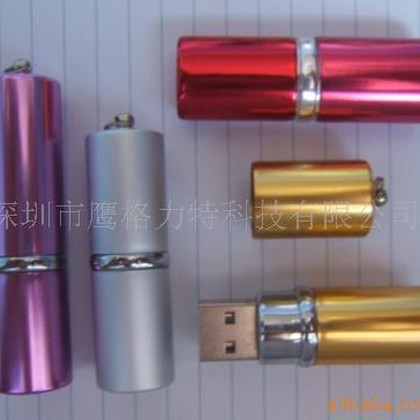}{USB stick}{lipstick}
  \teasersubfigure{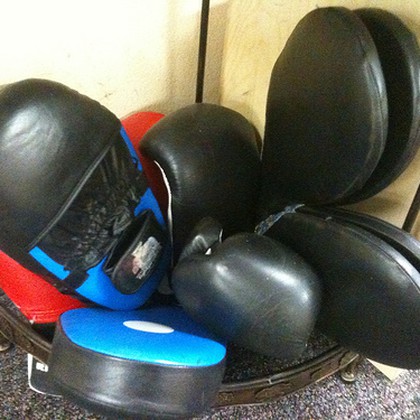}{boxing kit}{punching bag}
  \teasersubfigure{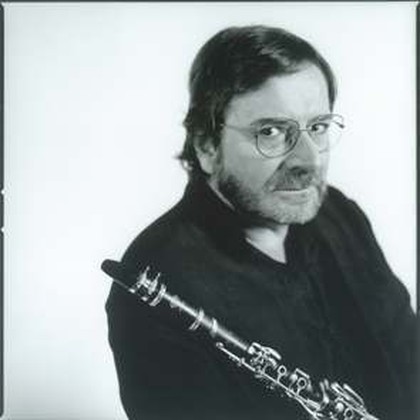}{clarinet}{oboe}
  \teasersubfigure{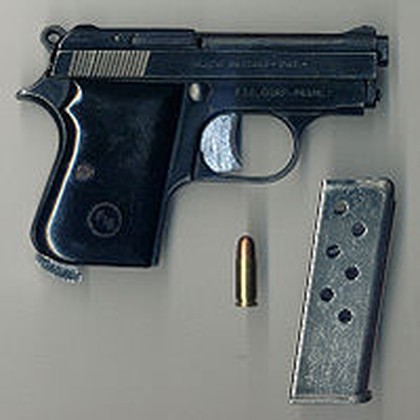}{pistol}{revolver}
  \teasersubfigure{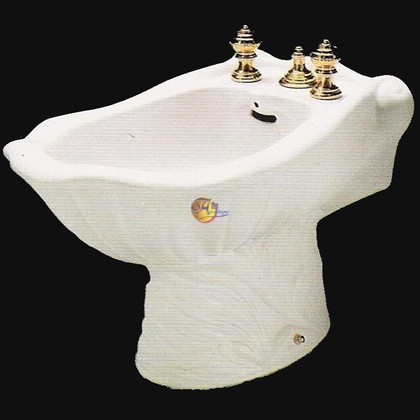}{bidet}{sink}
  \teasersubfigure{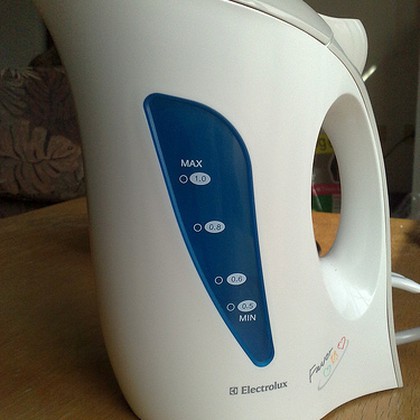}{electric kettle}{jug}
  \teasersubfigure{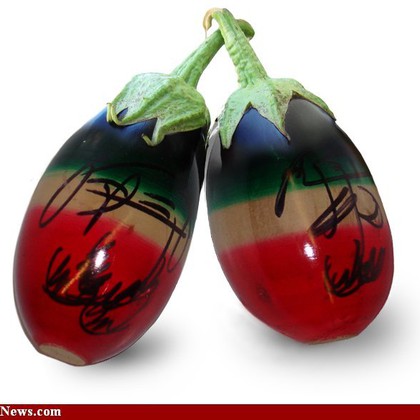}{painted eggplants}{maraca}
  \end{minipage}
  \caption{
    The \textbf{\bench{} dataset} -- 14 examples from the \nbench~OOD images.
     The new \textcolor{green!50!black}{human- and MLLM-verified OOD labels}, none in the ImageNet-1k class space, are in green.
     The \textcolor{red}{original ImageNet-1k labels} in red. More examples are in the supplementary material  
     (\cref{fig:supp-gallery}).
  }
  \label{fig:teaser}
\end{figure*}
Both in the OOD research and image classification domains one dataset stands out -- ImageNet \cite{deng2009imagenet},
the standard reference for visual recognition evaluation, from closed-set classification and segmentation to detection.
This carries over to OOD detection: the \imnet[1] validation set~\cite{russakovsky2015imagenet} is the in-distribution (ID) reference of the most popular large-scale OOD benchmarks \cite{bitterwolf2023ninco, imagenetood, huang2021mos, vaze2022ssb}.

These benchmarks adopt a similar approach to the construction of their OOD part. First, the set of OOD labels is defined, often arbitrarily,
and OOD images are then drawn from an image pool collected independently of the ID data. 
For example, ImageNet-O \cite{hendrycks2021imageneto}
collects  ImageNet-21k images from classes not inside ImageNet-1k,
for which ResNet50 \cite{he2016resnet} reports high confidence of an ImageNet-1k class, thus aiming at difficult examples. 
Alternatively, whole existing datasets can be used as OOD whenever their label space does not intersect the ID dataset labels. The disjoint label sets are more easily obtained for far-OOD datasets, but for near- or fine-grained OOD datasets the risk of label contamination is greatly increased.
For instance, Textures~\cite{cimpoi2014dtd}, collected for texture recognition, is used this way for both CIFAR~\cite{krizhevsky2009cifar} and \imnet[1] ID datasets.

The quality of data used for OOD benchmarking has also drawn attention.
NINCO \cite{bitterwolf2023ninco} inspected existing OOD benchmarks based on \imnet[1] and identified in-distribution samples in a subset of each.
\imnet[1], the source of in-distribution data, has been the subject of even more analyses:
Beyer \etal{} \cite{beyer2020imagenetreal} assessed multi-label annotations of the validation set and found many of the original labels at odds with careful human judgment.
Northcutt \etal{} \cite{northcutt2021labelerrors} estimated the level of label errors on several benchmarks and report $\sim$6\% on the ImageNet-1k validation set.

Recently, ReImageNet \cite{reimagenet} reannotated all 50,000 validation images with human experts and state-of-the-art LLMs.
We take advantage of this initiative
and introduce \bench{}, an OOD benchmark of \nood~images that, although originally labeled ID, depict concepts not represented in \imnet[1].
Its construction process follows a three-stage protocol (\cref{sec:construction}) built on per-image evidence: an open-vocabulary MLLM description, a reverse image search, the original ImageNet-1k label, and visual checks against the ImageNet-1k classes a VLM ranks closest to the image and those empirically confusable with its original label under ReImageNet's reannotation.
An image is admitted only when this evidence converges on a clearly identifiable object whose unambiguous fine-grained label lies \emph{outside} the 1000-class space.
The origin of the images -- they passed a human-based verification during ImageNet collection -- guarantees they are visually difficult to classify in the context of the 1000 ID classes, making them very suitable for OOD benchmarking.

Experiments conducted using a wide range of models \cite{fang2024eva02, zhai2023siglip} paired with OOD post-hoc methods \cite{mueller2025mahalanobispp, wang2022vim, ren2021rmaha} confirm that \bench{} is significantly more challenging than existing datasets~\cite{imagenetood, bitterwolf2023ninco, cimpoi2014dtd, vaze2022ssb} (see also \cref{fig:fpr_ours_vs_popular}).

\noindent In summary, our contributions are:
\begin{enumerate}
    \item \bench{}, an extremely challenging OOD benchmark of \nood{}~images from the inside of the \imnet[1] dataset.
    \item An extensive evaluation of supervised classifiers and zero-shot VLMs under both post-hoc scoring and VLM prompting (including negative-label prompting), which, as a side effect, confirms \bench{} difficulty.
    \item A novel approach for constructing an OOD dataset from label errors inside the ID data itself, verified per image by an MLLM description, reverse image search, and a VLM shortlist of the closest ID classes.
    \item Verified labels with fine-grained categorization, where applicable, for every \bench{} image, supporting future zero-shot and open-vocabulary OOD research beyond a binary ID/OOD (see \cref{fig:teaser} for examples).
\end{enumerate}

\section{Related Work}
\label{sec:related}

OOD detection and the study of \imnet[1] (IN-1k) label quality have developed apart: the first takes the validation labels as a clean in-distribution reference, the second documents how often those labels are wrong.
\bench{} sits at their intersection, turning verified label errors into near-boundary OOD samples.

\subsection{OOD detection methods}
\label{sec:related-detection}
\noindent\textbf{Post-hoc OOD detection.}
Most OOD detectors leave training untouched and score a pretrained classifier at test time.
They differ mainly in what intermediate outputs of the networks are utilized for OOD detection.
This divides them into four groups:
\emph{Logit-based} scores use the classifier output alone -- the maximum softmax
probability (MSP)~\cite{hendrycks2017msp} is the canonical baseline, and
MaxLogit and KL-Matching~\cite{hendrycks2022scaling}, Energy~\cite{liu2020energy}
and GEN~\cite{liu2023gen} refine the same readout.
\emph{Activation- and weight-based} scores first rectify the penultimate
activations or sparsify the classifier weights, then apply an energy-style
readout (ReAct~\cite{sun2021react}, DICE~\cite{sun2022dice},
ASH~\cite{djurisic2023ash}, SCALE~\cite{xu2024scale}).
\emph{Feature-based} scores flag inputs that fall far from the ID training
statistics: the Mahalanobis distance~\cite{lee2018mahalanobis} and its relative
variant~\cite{ren2021rmaha}, both recently strengthened by fitting on
$\ell_2$-normalized features (Mahalanobis++ and relative
Mahalanobis++~\cite{mueller2025mahalanobispp}); $k$-nearest
neighbors~\cite{sun2022knn}; similarity to the class-mean feature, plain cosine
or softmax-scaled, the class-centroid counterparts of the CLIP-side
scores~\cite{bitterwolf2023ninco}; and scores reading the decision geometry
directly (fDBD~\cite{liu2024fdbd}, NECO~\cite{ammar2024neco},
NAC~\cite{liu2024nac}).
\emph{Hybrid} scores combine features with logits (ViM~\cite{wang2022vim},
NNGuide~\cite{park2023nnguide}).
We evaluate all twenty-one of them on \bench{} (\cref{sec:supervised}).

\noindent\textbf{OOD detection with VLMs.}
CLIP-style models replace the supervised classification head with a dot product between the image embedding and a class-name embedding~\cite{radford2021clip}; OOD detectors then operate on these similarities rather than on classifier logits.
MCM~\cite{ming2022mcm} scores an image by its softmax-scaled similarity to the ID class-name embeddings alone.
Negative-label methods add a second, disjoint vocabulary: NegLabel~\cite{jiang2024neglabel} scores an image by the share of its text similarity that falls on the ID names rather than on a pool of negative labels (\cref{eq:neglabel}), so an image matching a negative more strongly than any ID name is flagged OOD.
Its negatives come from WordNet~\cite{miller1995wordnet}, a lexical database that groups English words into sets of synonyms (\emph{synsets}) linked by semantic relations such as hypernymy; NegLabel's NegMining step keeps the entries least similar to the ID class names, ensuring each negative refers to a concrete visual concept while remaining far from every ID class.
Later methods keep the scoring rule and revise the vocabulary.
EOE~\cite{cao2024eoe} replaces the lexicon with categories an LLM generates as plausible outliers for the given ID set and adds a penalty term to \cref{eq:neglabel} of NegLabel.
\subsection{Benchmarks and datasets}
\label{sec:related-benchmarks}
\noindent\textbf{OOD benchmarks for ImageNet.}
ImageNet-scale benchmarks take the \imnet[1] validation set as the ID reference and pair it with OOD images sourced from elsewhere.
OpenOOD~\cite{yang2022openood,zhang2023openoodv15} standardized the evaluation, gathering sets such as SSB~\cite{vaze2022ssb}, NINCO~\cite{bitterwolf2023ninco}, iNaturalist~\cite{huang2021mos}, and Textures~\cite{cimpoi2014dtd}. ImageNet-O~\cite{hendrycks2021imageneto} mines natural adversarial images. ImageNet-OOD~\cite{imagenetood} draws semantic-shift examples from ImageNet-21k.
NINCO and ImageNet-OOD additionally inspect every image to remove ID contamination shown to distort the popular sets~\cite{bitterwolf2023ninco}, yet both of them import out-of-distribution images from outside the \imnet[1] pipeline, potentially introducing a covariate shift that a detector can exploit.

\noindent\textbf{ImageNet label quality.}
A substantial body of work outside OOD detection shows that \imnet[1] labels are far from error-free.
Multi-label reannotation reveals that many original labels conflict with careful human judgment~\cite{beyer2020imagenetreal}, and the estimated error rate is high enough to change model rankings once corrected~\cite{northcutt2021labelerrors}.
Most recently, ReImageNet~\cite{reimagenet} reannotated all 50,000 validation images, flagging \nflagged{} as unassigned to any of the 1000 classes.
OOD detection has so far ignored these findings, treating the validation labels as a clean in-distribution reference.
\bench{} treats them as a useful signal, starting from ReImageNet's unassigned images and verifying, through the protocol of \cref{sec:construction}, which of them depict a genuine object from outside the IN-1k label space.

\begin{figure*}[t]
  \centering
  \includegraphics[width=\textwidth]{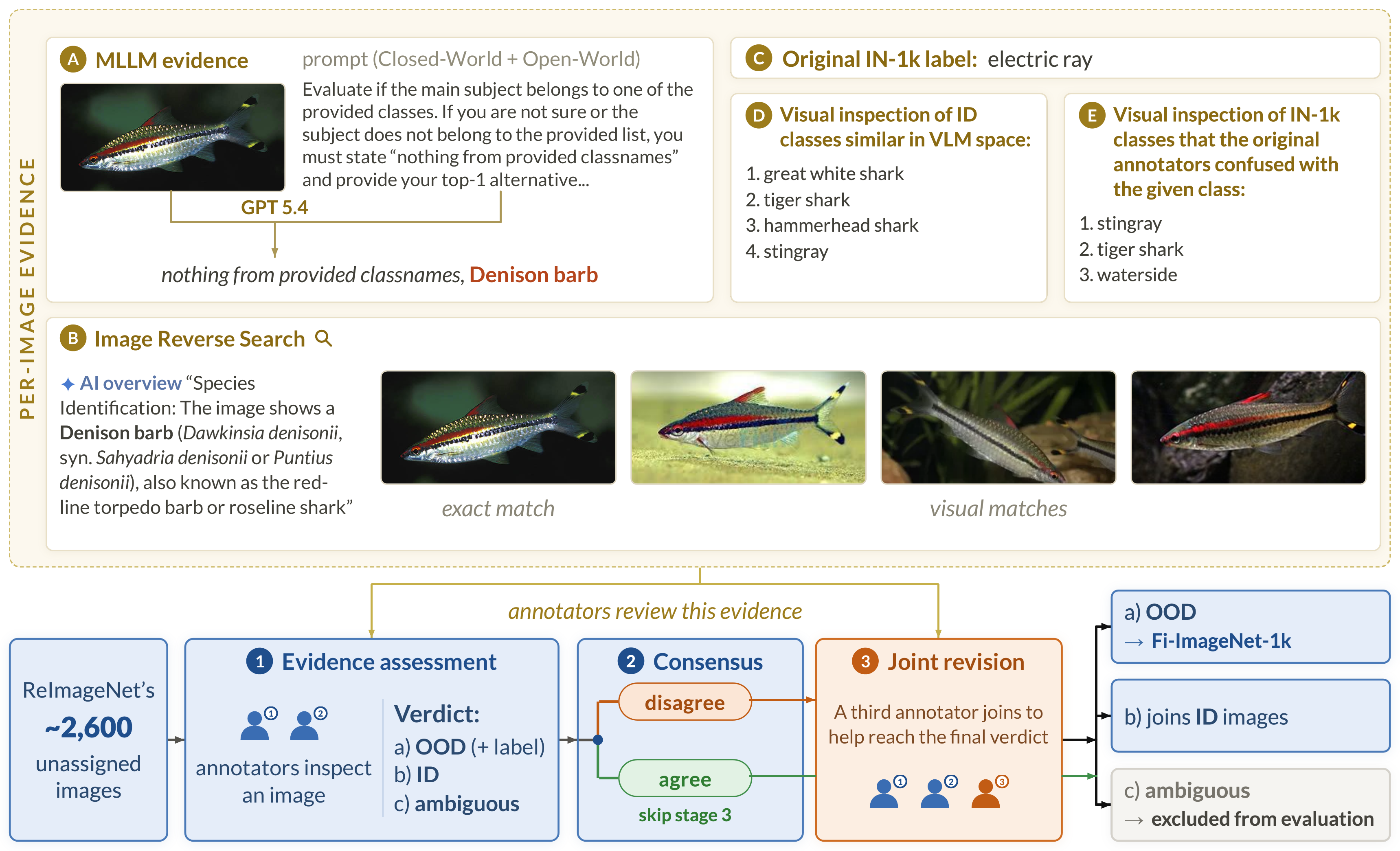}
  \caption{
  \textbf{The OOD annotation pipeline.} 
  \emph{Top:} the five sources of per-image evidence an annotator sees,
  shown here for a \emph{Denison barb} annotated originally as \emph{electric ray}.
  \emph{Bottom:} the three-stage protocol.
  \nood{}~images are admitted under consensus and form \bench; images resolved as ID amend in-distribution data, while ambiguous images are excluded from our experiments.
  }
  \label{fig:pipeline}
\end{figure*}

\section{Benchmark Construction}
\label{sec:construction}

We construct \bench{} from the ImageNet-1k validation images that ReImageNet reannotation~\cite{reimagenet} did not assign to any of the 1000~classes.
However, not being assigned to an ImageNet-1k class is a necessary, but not a sufficient condition for being considered an OOD.
In fact, a subset of the unassigned images is labeled as ``uncertain'', not as ``no ImageNet-1k class''.
We decided that, to declare an image as OOD, we have to be highly confident in its class (outside IN-1k).
The rest of the section describes the labeling effort of the ReImageNet rejected images,
the candidate pool, the evidence available to the annotators, and the three stages of the annotation process.

The proposed annotation pipeline is illustrated in \cref{fig:pipeline}.
In Stage~1, two annotators label every candidate independently based on the provided per-image evidence.
In Stage~2, agreements between the two annotators are accepted as final and skip Stage~3, while disagreements are passed to a third annotator for resolution.
Images are labeled only if the consensus among annotators is reached; otherwise, they are marked \emph{ambiguous}.

\noindent\textbf{Candidate pool.}
ReImageNet reannotated all 50{,}000 validation images, correcting numerous labeling errors \cite{reimagenet}.
In addition, their annotators were allowed to flag ambiguous images as ``uncertain'' or ``no ImageNet-1k class''.
However, they did not propose an alternative label for such images.
Therefore, we tasked our annotators with reviewing these samples and finding solid evidence for the decisions.

\noindent\textbf{Per-image evidence.}
Given an image, the annotators were provided with five independent sources of evidence (\cref{fig:pipeline}, top):
\begin{enumerate*}[label=(\Alph*)]
    \item an open-vocabulary description generated by GPT-5.4 through the OpenAI API (adapted from \cite{kisel2026multimodal}; the full prompt is given in \cref{supp:prompts} of the supplementary material), prompted with the possibility to output either one of the ImageNet-1k classes (closed-world) or an open-world prediction;
    \item reverse image search services, used to recover an exact match with provenance or to identify the most plausible fine-grained category;
    \item the original ImageNet-1k label, which indicates the ID class the image was originally confused with;
    \item the ImageNet-1k classes that a VLM ranks closest to the image, presented for visual comparison;
    \item a list of easily confusable ImageNet-1k classes, obtained by observing to which labels ReImageNet annotators often re-assigned images of the original class.
\end{enumerate*}

The last two sources address different failure modes.
(D) shows which ID classes the image resembles in feature space, regardless of coarse category -- a model-driven, image-level check.
(E) lists the ID classes that previous annotators frequently confused with the original label.
\Cref{fig:pipeline} shows an example of a \emph{Denison barb} annotated originally as \emph{electric ray} that must pass the comparison against classes from both (D) and (E) before it is admitted as OOD.

\noindent\textbf{Stage 1: evidence assessment.}
Two annotators, with prior large-scale annotation experience and acquired knowledge of the full ImageNet-1k label space, labeled all images independently, issuing one of three verdicts:
\begin{enumerate*}[label=(\alph*)]
   \item ~\emph{OOD}: the depicted object lies outside the 1000-class space \emph{and} no object of any ImageNet-1k class is visible anywhere in the image; the annotator proposes an OOD label;
   \item ~\emph{ID}: the image contains a valid ImageNet-1k object;%
   \item ~\emph{ambiguous}: an object is visible but cannot be resolved to a single label with the available evidence; or the image is corrupted, object-free, or too degraded to judge.
\end{enumerate*}

Furthermore, the annotators were also asked to flag an image as \emph{fine-grained} when the depicted object is a close sibling of its original ImageNet-1k class, \eg, a different breed, a congeneric species, or a near-identical instrument.

\noindent\textbf{Stage 2: consensus.}
The two independent verdicts are compared.
When the annotators agree, the verdict is final and the image skips Stage~3: agreed \emph{OOD} images are admitted, agreed \emph{ID} images are returned to the ID data, and agreed \emph{ambiguous} images are excluded from our experiments.
Only the images on which the two annotators disagree are promoted to Stage~3.

\noindent\textbf{Stage 3: joint revision.}
A third annotator with the same training joins the first two, and all three review the disputed images together.
An image receives a verdict only if the three annotators reach full consensus; if they do not, the image is marked \emph{ambiguous} and excluded.

\noindent\textbf{Outcome.}
The protocol verified \nflagged{}~images, of which \nnosure{} were marked by ReImageNet as ``uncertain'' and \nempty{} as ``no ImageNet-1k class''.
Of all images, \nbench{} (25.2\%) were identified as OOD with a verified label.
They cover \nbenchlab{}~distinct concepts and are further divided into 358 fine-grained instances (54.7\%) and 297 ordinary.
Next, \npoolid{} (37.0\%) were found to depict a concept from ReImageNet, returned to the ID pool and subsequently reported to the authors.
For the rest, \nambig{} (37.7\%), our annotators could not commit to a label, and such images were omitted from our experiments.

Analyzing the two subsets (``no ImageNet-1k class'' vs. ``uncertain'') of the candidate pool:
the \nempty{} images initially declared outside ImageNet split into 205 (43.2\%) certified OOD, 197 (41.6\%) containing a valid ImageNet-1k object, and 72 (15.2\%) remaining ambiguous.
Of the \nnosure{} ``uncertain'' images, just 450 (21.2\%) were certified OOD, 765 (36.0\%) were resolved to an ImageNet-1k class, and 908 (42.8\%) remained ambiguous -- the ``uncertain'' flag was thus weaker evidence of OOD content.
In contrast, the fine-grained share of the certified OOD images is nearly the same between the two flags: 117/205 (57.1\%) for outside IN-1k, 241/450 (53.6\%) for ``uncertain''.

\section{Experiments}
\label{sec:experiments}

We evaluated \bench{} using two main methodologies -- supervised models trained or fine-tuned on \imnet[1] and zero-shot VLMs.
Each model is evaluated with a set of post-hoc OOD detectors.
We first test whether the benchmark remains difficult under varying backbone architectures and OOD detectors for both methodologies (\cref{sec:supervised,sec:vlms}).
Next, we extend the OOD detection suite for the VLMs with methods based on negative labels to isolate the role of the textual vocabulary (\cref{sec:neg-labels}), compare \bench{} with existing ImageNet-scale OOD benchmarks (\cref{sec:comparison}), and quantify the effect of the number of test images (\cref{sec:dataset_size}).

\subsection{Evaluation Setup}
\label{sec:setup}

\noindent\textbf{Models.}
The supervised family consists of eight public backbones fine-tuned on \imnet[1] for a classification task.
The backbones are chosen for diversity, and they differ in the supervision and pre-training regime they receive beyond \imnet[1] fine-tuning, since a larger pre-training corpus is known to improve OOD detection~\cite{bitterwolf2023ninco}.
ResNet50~\cite{he2016resnet} is trained on \imnet[1] labels only.
ViT-B/16-384~\cite{dosovitskiy2021vit}, ConvNeXt-B~\cite{liu2022convnext} and SwinV2-B-256~\cite{liu2022swinv2} are pre-trained on \imnet[21] labels and fine-tuned on \imnet[1].
ConvNeXtV2-L~\cite{woo2023convnextv2} and EVA02-L-448~\cite{fang2024eva02} use self-supervision for pre-training (FCMAE~\cite{woo2023convnextv2} via masked image modeling~\cite{bao2022beit,he2022mae}), then fine-tuned on \imnet[21] and \imnet[1].
DINOv2-g/14~\cite{oquab2023dinov2} and DINOv3-7B~\cite{simeoni2025dinov3} are also pre-trained in a self-supervised manner on LVD-142M~\cite{oquab2023dinov2} and LVD-1689M~\cite{simeoni2025dinov3}, respectively, and kept frozen; we train only a linear probe on the \imnet[1] training split.

The zero-shot family consists of eight text-image contrastive VLMs without any ImageNet-specific adaptation or fine-tuning.
Models, their sizes and \imnet[1] top-1 accuracies are listed in \cref{tab:supervised,tab:vlm-posthoc}.

\noindent\textbf{OOD Detectors.}
For the supervised methods, we combine them with the 21 post-hoc OOD detectors of \cref{sec:related-detection}, covering four groups:
i) five logit-based (MSP, MaxLogit, KL-Matching, Energy, GEN),
ii) four activation- and weight-based (Energy+ReAct, ASH-S, SCALE, DICE),
iii) ten feature-based (Mahalanobis, relative Mahalanobis, Mahalanobis++, relative Mahalanobis++, $k$-NN, cosine, MCM, fDBD, NECO, NAC),
and iv) two hybrid (ViM, NNGuide).

For the VLMs, we consider two detector types: logit-based and vocabulary-based.
Let $h(x)$ be the $\ell_2$-normalized embedding of an image $x$ and $t$ the $\ell_2$-normalized embedding of a text label, both produced by the VLM in its shared image-text space. 
A label embedding is the average over the eight prompt templates of~\cite{radford2021clip, volkov2025vlmembeddings}, listed in \cref{supp:prompts} of the supplementary material.
Every VLM score used in this paper is built from the temperature-scaled, exponentiated cosine similarity
\begin{equation}
a(x,t)=\exp\!\left(h(x)^\top t/\tau\right),
\label{eq:affinity}
\end{equation}
which plays the role of the exponentiated logit of a supervised classifier. Following NegLabel~\cite{jiang2024neglabel}, we use the temperature $\tau{=}0.01$.
Let $\{t_c\}_{c=1}^{C}$ be a set of embeddings computed for the $C{=}1000$ ID class names, the standard zero-shot classifier assigns to $x$ the softmax posterior
\begin{equation}
p(c\mid x)=\frac{a(x,t_c)}{\displaystyle\sum\nolimits_{c=1}^{C} a(x,t_c)}.
\label{eq:zeroshot}
\end{equation}

The \emph{logit-based} detectors are the usual post-hoc scores of a classifier, computed from the zero-shot logits $h(x)^\top t_c/\tau$ or from the posterior of \cref{eq:zeroshot}: MCM~\cite{ming2022mcm}, which is $\max_c p(c\mid x)$, MaxLogit, Energy, GEN, Entropy and Margin all operate on image-level logits, while GL-MCM~\cite{miyai2025glmcm} incorporates into MCM the maximum softmax over the patch-level logits.

\emph{Vocabulary-based} detectors -- NegLabel~\cite{jiang2024neglabel}, EOE~\cite{cao2024eoe}, and our variant of NegLabel with LLM-generated negatives adopted from EOE -- extend the ID label set with $M$ negative (OOD) labels with corresponding embeddings $\{\hat{t}_m\}_{m=1}^{M}$, whose similarities to $x$ are added only to the denominator.
An image is then scored by the share of its total similarity that falls on the ID names,
\begin{equation}
s(x)=
\frac{\displaystyle\sum\nolimits_{c=1}^{C} a(x,t_c)}
     {\displaystyle\sum\nolimits_{c=1}^{C} a(x,t_c)
      +\displaystyle\sum\nolimits_{m=1}^{M} a(x,\hat{t}_m)},
\label{eq:neglabel}
\end{equation}
so that images more similar to negative labels have a lower score $s(x)$ and are flagged as OOD.

\noindent\textbf{Data.}
In all experiments, we evaluate the described OOD benchmarks against the same in-distribution dataset, consisting of \nid{}~\imnet[1] validation images, which are not in \bench{} and were not flagged as ambiguous (\cref{sec:construction}).

\noindent\textbf{Metrics.}
Detectors assign a scalar to all images, where a larger value indicates an in-distribution sample.
We follow the standard evaluation metrics for the OOD detection task, specifically:
i) false positive rate at 95\% true positive rate (FPR@95), which is the fraction of OOD images accepted as ID at the threshold that correctly classifies 95\% of the ID data;
and ii) area under the receiver operating characteristic curve (AUROC), which is a threshold-free metric and measures ranking quality over all operating points.

\subsection{Post-hoc Supervised OOD Detection}
\label{sec:supervised}

\newcommand{\grouphead}[1]{%
  \multicolumn{7}{@{}l@{}}{\makebox[0pt][l]{\makebox[\columnwidth][c]{\textit{#1}}}}}

\begin{table}[t]
\centering
\small
\setlength{\tabcolsep}{0pt}
\renewcommand{\arraystretch}{1.05}
\caption{
    \textbf{Post-hoc OOD detection on \bench{}.}
    For every model, the best detector (out of the 21~post-hoc detectors for the supervised models and the 7~logit-based detectors for the zero-shot VLMs, \cref{sec:setup}) in terms of FPR@95 and AUROC is shown in gray.
    Top-1 -- classification accuracy on the reannotated \imnet[1] validation set (ReImageNet~\cite{reimagenet}); for the VLMs it is the zero-shot accuracy obtained with the text labels of the \imnet[1] classes.
    Linear probes on a frozen backbone are marked by $\dagger$.
    \textbf{Bold:} best column value within each group.
}
\label{tab:supervised}
\label{tab:main-results}
\label{tab:all-models-inside-in1k}
\label{tab:vlm-posthoc}
\begin{tabular*}{\columnwidth}{
    @{}
    l
    @{\extracolsep{\fill}}
    r
    c
    r
    @{\hspace{1pt}}
    l
    r
    @{\hspace{1pt}}
    l
    @{}
}
\toprule
Model & Params & Top-1
& \multicolumn{2}{c}{FPR@95 $\downarrow$}
& \multicolumn{2}{c@{}}{AUROC $\uparrow$} \\
\midrule
\grouphead{Supervised pre-trained models} \\
\addlinespace[1pt]
ResNet50              &  25.6M & 84.9          & 80.2          & \g{rMaha++} & 74.2          & \g{rMaha++} \\
SwinV2-B-256          &  87.9M & 89.2          & 73.3          & \g{NAC}     & 77.9          & \g{rMaha++} \\
ConvNeXt-B            &  88.6M & 89.1          & 69.2          & \g{rMaha++} & 80.3          & \g{rMaha++} \\
ViT-B/16-384          &  86.9M & 89.3          & 69.0          & \g{Maha++}  & 81.4          & \g{Maha}    \\
DINOv2-g/14 $\dagger$ &   1.1B & 89.1          & 68.7          & \g{ViM}     & 77.5          & \g{GEN}     \\
ConvNeXtV2-L          & 198.0M & 90.3          & 67.8          & \g{rMaha++} & 83.1          & \g{Maha++}  \\
EVA02-L-448           & 305.1M & \textbf{91.5} & 65.0          & \g{Maha++}  & \textbf{84.2} & \g{rMaha++} \\
DINOv3-7B $\dagger$   &   6.7B & 90.1          & \textbf{64.4} & \g{rMaha++} & 81.0          & \g{MCM}     \\
\midrule
\grouphead{Zero-shot vision--language models} \\
\addlinespace[1pt]
CLIP-B/16             & 149.6M & 73.5          & 87.0          & \g{GL-MCM}  & 69.1          & \g{GL-MCM}  \\
CLIP-L/14             & 427.9M & 81.3          & 84.0          & \g{GL-MCM}  & 72.7          & \g{Entropy} \\
AIMv2-L               & 436.7M & 83.2          & 82.9          & \g{Entropy} & 76.6          & \g{Entropy} \\
MetaCLIP2-H/14        &   1.9B & 86.6          & 76.3          & \g{GL-MCM}  & 79.8          & \g{GL-MCM}  \\
PE-Core-L/14          & 671.1M & 88.4          & 76.2          & \g{Entropy} & 80.0          & \g{Entropy} \\
OpenCLIP-H/14         & 986.7M & 88.5          & 72.5          & \g{Entropy} & \textbf{81.1} & \g{GEN}     \\
SigLIP-SO400M         & 878.0M & 87.8          & \textbf{71.8} & \g{GL-MCM}  & 80.3          & \g{Entropy} \\
SigLIP2-g             &   1.9B & \textbf{88.6} & \textbf{71.8} & \g{Entropy} & 81.0          & \g{Entropy} \\
\bottomrule
\end{tabular*}
\end{table}

We observe (\cref{tab:supervised}) that at 95\% true positive rate, every supervised configuration accepts most \bench{} images as ID.
The best result is 64.4\% FPR@95, achieved by the DINOv3-7B linear probe with relative Mahalanobis++; and the worst reached 80.2\% (for ResNet50).
Overall, the Mahalanobis-family score is the best OOD detector for six of the eight backbone models, all six with Mahalanobis++ or its relative variant~\cite{mueller2025mahalanobispp}, yet the gain over the second-best OOD detector is marginal -- 2.1 points (ResNet50) and 2.8 points for DINOv3-7B.
The complete per-detector breakdown is available in \cref{tab:supp-all-detectors} of the supplementary material.
Neither a stronger backbone nor a better post-hoc method can sufficiently address the challenge of \bench{}.

\subsection{Zero-Shot VLMs as OOD Detectors}
\label{sec:vlms}

The results for the seven logit-based OOD detectors are provided in \cref{tab:vlm-posthoc}.
The best FPR@95 ranges from 87.0\% (CLIP-B/16) to 71.8\% (SigLIP2-g and SigLIP-SO400M). 
GL-MCM, the only detector in the pool that reads patch-level similarities in addition to the pooled image embedding, is the best detector for four of the eight VLMs, suggesting that local evidence carries an OOD signal the global embedding misses.

WordNet-based NegLabel (\cref{tab:neg-sources}, column WN) improves on the best logit-space OOD detector for six of the eight VLMs, by 1.4 (PE-Core-L/14, OpenCLIP-H/14) to 8.7 points (CLIP-L/14).
This experiment suggests that the size and selection of the negative (OOD) textual vocabulary is the crucial aspect for OOD performance in VLMs, and we provide further analysis below.

\subsection{Zero-Shot with Negative-Label Vocabularies}
\label{sec:neg-labels}

We incorporate the negative (OOD) vocabulary into the zero-shot methods using the NegLabel~\cite{jiang2024neglabel} score of \cref{eq:neglabel}.
With the model and the ID names fixed, the negative vocabulary $\{\hat{t}_j\}_{j=1}^{M}$ is the only free component of \cref{eq:neglabel}; this section analyzes its impact.

\begin{table}[t]
\centering
\small
\setlength{\tabcolsep}{0pt}
\renewcommand{\arraystretch}{1.05}
\caption{
    \textbf{VLMs with negative vocabularies on \bench{}.}
    FPR@95 (\%, $\downarrow$) per VLM; the columns differ only in the negative vocabulary of \cref{sec:neg-labels}.
    \emph{logit-sp.} is the best of the seven logit-space detectors of \cref{tab:vlm-posthoc} (column copied from \cref{tab:supervised}).
    \emph{EOE}~\cite{cao2024eoe}: LLM-generated outlier categories.
    NegLabel~\cite{jiang2024neglabel} with four vocabularies: its NegMined WordNet lexicon (\emph{WN}, $M{=}5000$), our LLM-generated near siblings of every ID class (\emph{GPT}), the union of \emph{WN} and \emph{GPT} lexicons (\emph{WN+GPT}), and the \nbenchlab{} distinct OOD labels of \bench{} itself (\emph{oracle}), a lower bound on FPR@95 rather than a detector.
    EOE and GPT are means over 30 and 10 generations of GPT API, respectively.
    \textbf{Bold:} best deployable vocabulary per row.
}
\label{tab:neg-sources}
\label{tab:neg-labels}
\begin{tabular*}{\columnwidth}{@{}l@{\extracolsep{\fill}}clcrcr@{}}
\toprule
& & & \multicolumn{4}{c@{}}{NegLabel} \\
\cmidrule{4-7}
Model & logit-sp. & \multicolumn{1}{l}{EOE} & \multicolumn{1}{c}{WN} & \multicolumn{1}{c}{GPT} & \multicolumn{1}{c}{WN+GPT} & oracle \\
\midrule
CLIP-B/16 \cite{radford2021clip}          & 87.0 & 79.7 & 80.2 & 73.4 & \textbf{71.7} & 33.6 \\
CLIP-L/14 \cite{radford2021clip}          & 84.0 & 74.7 & 75.3 & 63.3 & \textbf{62.9} & 21.5 \\
AIMv2-L \cite{fini2024aimv2}              & 82.9 & 72.2 & 75.1 & \textbf{60.6} & 62.1 & 15.1 \\
PE-Core-L/14 \cite{bolya2025pe}           & 76.2 & 66.7 & 74.8 & \textbf{55.5} & \textbf{55.5} & 8.2 \\
MetaCLIP2-H/14 \cite{chuang2025metaclip2} & 76.3 & 69.0 & 77.3 & \textbf{55.1} & 56.8 & 9.2 \\
OpenCLIP-H/14 \cite{cherti2023openclip}   & 72.5 & 68.2 & 71.1 & \textbf{53.6} & 54.4 & 8.7 \\
SigLIP-SO400M \cite{zhai2023siglip}       & 71.8 & 66.6 & 74.0 & \textbf{55.6} & 55.8 & 7.8 \\
SigLIP2-g \cite{tschannen2025siglip2}     & 71.8 & 64.6 & 68.9 & 51.3 & \textbf{51.2} & 6.0 \\
\midrule
Mean           & 77.8 & 70.2 & 74.6 & \textbf{58.6} & 58.8 & 13.8 \\
\bottomrule
\end{tabular*}
\vspace{-2mm}
\end{table}

We compare four vocabulary construction techniques applied to zero-shot VLMs' output.
The comparison is shown in \cref{tab:neg-labels}.
\emph{EOE} \cite{cao2024eoe} performs the construction by asking an LLM to generate plausible outlier categories for the ID set and adds a penalty term to \cref{eq:neglabel}.
In our evaluation, we follow the default setting from the EOE paper \cite{cao2024eoe} and use 500 categories with a fine-grained prompt, the strongest of the three prompts on \bench{} (for performance using all three prompts proposed by EOE see \cref{tab:supp-eoe-prompts} of the supplementary material). The remaining methods use NegLabel with different vocabulary pools.

\emph{WN} is the original WordNet lexicon \cite{miller1995wordnet}, comprising 79{,}403 head lemmas of all noun and adjective synsets, reduced by NegMining to $M$~candidates.
We use $M{=}5000$, which beats the default ($M{=}10{,}000$) for seven of the eight VLMs on \bench{} (see \cref{fig:supp-wordnet} of the supplementary material for the ablation on $M$).

\emph{GPT}, our variant, replaces both the lexicon and NegMining: we prompt the LLM once per ID class to generate 20 fine-grained near siblings for that class, discard any sibling that is itself an ID class name, and concatenate the per-class lists into a single pool of unique $\approx$19{,}600 labels.

\emph{WN+GPT} is the union of the WordNet lexicon, representing far-OOD concepts, and negative vocabulary obtained from the GPT method described above, representing near-fine-grained concepts; $\approx$24,600 labels in total.

The \emph{oracle} forms the vocabulary from the correct OOD labels of \bench{}.
It serves as our empirical lower bound for FPR@95 rather than as a detector, and it tests whether the VLM can separate the images once the right labels are available.

\Cref{tab:neg-sources} supports four observations.
First, every negative vocabulary improves on the best logit-space detector (\emph{logit-sp.}, copied from \cref{tab:vlm-posthoc}), by 3.2 points (WN) to 19.2 points (GPT) on average, demonstrating the importance of negative vocabulary modeling.
Next, GPT-mined negative labels beat both WordNet and EOE for \emph{every} tested VLM and reduce the mean from 74.6\% (WN) to 58.6\%, by 6.8 points on CLIP-B/16 up to 22.2 on MetaCLIP2-H/14.
EOE is second for every VLM (mean 70.2\%), but only with its fine-grained prompt; the other two fall to the MCM level.

What separates the GPT-mined from WordNet is where the negatives come from, not their amount: as WordNet is a general-purpose lexicon whose entries are selected only for being \emph{far} from every ID name, it is unlikely to contain the near sibling that a given \bench{} image actually depicts.
Prompting per ID class produces those near siblings directly, which seems to be the superior approach for our benchmark.

Combining the two vocabularies confirms this: the union \emph{WN+GPT} changes the mean by 0.2 points over GPT alone (58.8\% vs.\ 58.6\%) and helps only three of the eight VLMs.
Therefore, WordNet's contribution on \bench{} is limited once fine-grained negatives per class (\eg, mined with GPT) are present.
\Cref{tab:supp-popular-vocab} of the supplementary material presents the results of all four vocabularies on other OOD datasets, where the combination of WN and GPT outperforms its parts for several datasets, most notably NINCO and ImageNet-OOD.

The oracle reduces the mean FPR@95 to 13.8\%, with five VLMs below 10\% -- given the correct negative vocabulary, the feature representation of current VLMs is sufficient to solve \bench{}.
The limiting factor is thus vocabulary coverage: the 45-point gap between the best deployable pool (58.6\%) and the oracle shows that current methods for mining, or otherwise modeling, the OOD vocabulary remain far behind, and automatic construction of negative prompts is still an unresolved challenge.

\subsection{Comparison to Existing OOD Benchmarks}
\label{sec:comparison}

\begin{table}[t]
\centering
\small
\setlength{\tabcolsep}{0pt}
\renewcommand{\arraystretch}{1.05}
\caption{
    \textbf{\bench{} vs.\ popular OOD benchmarks.}
    FPR@95 (\%, $\downarrow$) and AUROC (\%, $\uparrow$) of EVA02-L-448 with the best of the 21 post-hoc detectors per dataset, named in gray; $n$ is the number of OOD images. CLEAN-OOD is the union of the cleaned subsets of the 11 datasets above it~\cite{bitterwolf2023ninco}. \bench{} is the most difficult of the 14 datasets.
}
\label{tab:eva-siglip-subsets}
\begin{tabular*}{\columnwidth}{
    @{}
    l
    @{\extracolsep{\fill}}
    r
    r
    @{\hspace{2pt}}
    l
    r
    @{\hspace{2pt}}
    l
    @{}
}
\toprule
Dataset & $n$
& \multicolumn{2}{c}{FPR@95 $\downarrow$}
& \multicolumn{2}{c@{}}{AUROC $\uparrow$} \\
\midrule
iNat-OOD Plants~\cite{huang2021mos}& 10\,000&  1.2 & \g{ViM}      & 99.8 & \g{ViM}    \\
OpenImage-O~\cite{wang2022vim}& 17\,632&  5.8 & \g{Maha++}   & 98.8 & \g{Maha++} \\
360OpenSet~\cite{bendale2016openset}& 11\,800& 11.9 & \g{E+ReAct}  & 97.0 & \g{Maha++} \\
COOD~\cite{galil2023cood}& 10\,350& 13.8 & \g{Maha++}   & 96.6 & \g{Maha++} \\
ImageNet-O~\cite{hendrycks2021imageneto}&  2\,000& 18.0 & \g{Maha++}   & 95.8 & \g{Maha++} \\
Textures~\cite{cimpoi2014dtd}&  5\,640& 21.2 & \g{Maha++}   & 94.1 & \g{Maha++} \\
Places~\cite{huang2021mos}& 10\,000& 25.3 & \g{NNGuide}  & 94.6 & \g{ViM}    \\
SSB easy~\cite{vaze2022ssb}& 50\,000& 30.0 & \g{NECO}     & 92.1 & \g{Maha++} \\
SSB hard~\cite{vaze2022ssb}& 49\,000& 33.1 & \g{ViM}      & 92.7 & \g{ViM}    \\
Species~\cite{hendrycks2022scaling}& 10\,000& 40.2 & \g{Maha}     & 88.4 & \g{ViM}    \\
IN1k-OOD~\cite{wang2022pascl}& 50\,000& 48.2 & \g{NAC}      & \textbf{83.3} & \g{GEN}    \\
\midrule
CLEAN-OOD~\cite{bitterwolf2023ninco}&  2\,715& 12.2 & \g{Maha++}   & 97.4 & \g{Maha++} \\
NINCO~\cite{bitterwolf2023ninco}&  5\,879& 16.9 & \g{Maha++}   & 96.7 & \g{Maha++} \\
ImageNet-OOD~\cite{imagenetood}& 31\,802& 18.8 & \g{NECO}     & 96.2 & \g{NECO}   \\
\midrule
\textbf{\bench{}} & \nbench{} & \textbf{65.0} & \g{Maha++} & 84.2 & \g{rMaha++} \\
\bottomrule
\end{tabular*}
\vspace{-2mm}
\end{table}

\Cref{tab:eva-siglip-subsets} compares \bench{} and ImageNet-scale OOD benchmarks of \cref{sec:related-benchmarks} using EVA02-L-448, the supervised backbone with the best accuracy on ReImageNet (\cref{tab:supervised}), and the same pool of 21 OOD detectors.
The first 11~rows report each dataset in full size.
However, these are known to contain images of \imnet[1] classes~\cite{bitterwolf2023ninco}. Thus, a detector that correctly accepts such an image is penalized, inflating its FPR@95.
We therefore report three benchmarks in which this contamination has been removed.

CLEAN-OOD is our aggregation of cleaned data from subsets of standard OOD datasets that the NINCO team annotated to quantify the contamination of those datasets with ImageNet-1k classes.
We union the 11 surviving sub-samples into a single set of 2715 images.
NINCO~\cite{bitterwolf2023ninco} and ImageNet-OOD~\cite{imagenetood} are the two published benchmarks whose construction explicitly targets ID contamination, making them the most directly comparable baselines.

The best FPR@95 on the 11 full OOD datasets ranges from 1.2\% to 48.2\%; \bench{} reaches 65.0\%, 16.8~points above the hardest of them.
The gap to the curated benchmarks is larger: CLEAN-OOD, NINCO and ImageNet-OOD reach 12.2\%, 16.9\% and 18.8\%, respectively; \ie, \bench{} has 3.8\texttimes{} the FPR@95 of NINCO and 3.5\texttimes{} that of ImageNet-OOD.
Its AUROC is comparable to the most difficult IN1k-OOD dataset (within a percentage point), a much smaller margin than in FPR@95.
Since we report the best achieved results for each dataset, the performance gap is not attributable to a single method with unfavorable performance on \bench{}.
Per-model results on the full datasets are in \cref{tab:supp-per-model-datasets} of the supplementary material.

\subsection{Effect of the OOD Dataset Size}
\label{sec:dataset_size}

The compared OOD dataset sizes range from \nbench{}~images in \bench{} to tens of thousands.
Dataset size affects the accuracy of the FPR@95 estimate, but not its expected value.
For a fixed model--detector pair, let $s(x)$ be the ID score and $t_{95}$ the threshold that accepts 95\% of the ID set.
Given an OOD dataset $D=\{x_i\}_{i=1}^{n}$, let
\begin{equation}
    Z_i=\begin{cases}
      1, & \text{if } s(x_i)\geq t_{95},\\
      0, & \text{otherwise,}
   \end{cases}
\end{equation}
indicate that $x_i$ is wrongly accepted as ID; the empirical FPR@95 is then the sample mean $\widehat{p}_D=\frac{1}{n}\sum_{i=1}^{n}Z_i$.
For image-level sampling, $Z_i\sim\mathrm{Bernoulli}(p_D)$, where $p_D$ is the population FPR of dataset $D$.
Conditional on the ID threshold, the expectation and the variance of the sample mean over $n$ images are
\begin{equation}
    \mathbb{E}\!\left[\widehat{p}_D\right] = p_D,
    \qquad
    \mathrm{Var}\!\left[\widehat{p}_D\right]
    =
    \frac{p_D(1-p_D)}{n}.
    \label{eq:fpr_variance}
\end{equation}
Hence, $\widehat{p}_D$ is unbiased for every $n$, and since $p_D(1-p_D)\leq 0.25$, its standard error is at most $1/(2\sqrt{n})$.
For $n=\nbench{}$ this bound is 1.95 points, giving a 95\% confidence half-width for $\widehat{p}_D$ below 3.9 points.
This is far below the 16.8-point gap between \bench{} and the hardest full dataset in \cref{tab:eva-siglip-subsets} (12.0 points for the fixed EVA02-L-448 and relative Mahalanobis++ pair of \cref{fig:fpr_ours_vs_popular}).

In \cref{supp:size} of the supplementary material, we confirm this by re-evaluating every compared dataset on 1000~random \nbench{}-image subsets of itself: the subset mean reproduces the full-set FPR@95 to within 0.13 points, while the sampling standard deviation is only 1.39–1.80 points, compared with a 12.7–16.4-point spread across datasets (\cref{tab:supp-size}).

\section{Conclusions}
\label{sec:conclusions}

We introduced \bench{}, an OOD benchmark.
It consists of images that ReImageNet did not assign to any of the 1000 classes. The images were validated by
a three-stage process combining human judgment with evidence from an open-vocabulary MLLM, reverse image search, the original label, and the ID classes closest to the image.
The result is \nbench{}~OOD images from \nbenchlab{}~classes, each with a verified label.

The construction is a general recipe for building OOD benchmarks from images
of established ID datasets with label errors: 
collect the unassigned or mislabeled images, gather complementary evidence about their content, and compare them against their most plausible ID classes.

\bench{} is more challenging than the commonly used OOD benchmarks: 
neither combination of eight backbones paired with 21 post-hoc detectors nor eight zero-shot VLMs achieved FPR@95 below 51\%.
Even when selecting the best-performing detector for each standard OOD dataset, including for ours \bench{}, the performance gap is 16.8 percentage points above IN1k-OOD, the hardest of the 14 prior OOD datasets (\cref{fig:fpr_ours_vs_popular}).

Given the benchmark labels as negatives, NegLabel paired with SigLIP2-g reaches FPR@95 of 6.0\%: the representation suffices, and the negative vocabulary is the bottleneck.
LLM-generated per-class siblings, the best deployable vocabulary, reach 51.3\%; closing the remaining gap to the oracle with an automatically constructed vocabulary is an open problem that \bench{} makes measurable.

\newpage
{
    \small
    \bibliographystyle{ieeenat_fullname}
    \bibliography{refs}

\begin{thebibliography}{59}
\providecommand{\natexlab}[1]{#1}
\providecommand{\url}[1]{\texttt{#1}}
\expandafter\ifx\csname urlstyle\endcsname\relax
  \providecommand{\doi}[1]{doi: #1}\else
  \providecommand{\doi}{doi: \begingroup \urlstyle{rm}\Url}\fi

\bibitem[Ammar et~al.(2024)Ammar, Belkhir, Popescu, Manzanera, and
  Franchi]{ammar2024neco}
Mou{\"i}n~Ben Ammar, Nacim Belkhir, Sebastian Popescu, Antoine Manzanera, and
  Gianni Franchi.
\newblock {NECO}: Neural collapse based out-of-distribution detection.
\newblock In \emph{International Conference on Learning Representations
  (ICLR)}, 2024.

\bibitem[Bao et~al.(2022)Bao, Dong, Piao, and Wei]{bao2022beit}
Hangbo Bao, Li Dong, Songhao Piao, and Furu Wei.
\newblock {BEiT}: {BERT} pre-training of image transformers.
\newblock In \emph{International Conference on Learning Representations
  (ICLR)}, 2022.

\bibitem[Bendale and Boult(2016)]{bendale2016openset}
Abhijit Bendale and Terrance~E. Boult.
\newblock Towards open set deep networks.
\newblock In \emph{IEEE Conference on Computer Vision and Pattern Recognition
  (CVPR)}, 2016.

\bibitem[Beyer et~al.(2020)Beyer, H{\'e}naff, Kolesnikov, Zhai, and van~den
  Oord]{beyer2020imagenetreal}
Lucas Beyer, Olivier~J. H{\'e}naff, Alexander Kolesnikov, Xiaohua Zhai, and
  A{\"a}ron van~den Oord.
\newblock Are we done with {ImageNet}?
\newblock \emph{arXiv preprint arXiv:2006.07159}, 2020.

\bibitem[Bitterwolf et~al.(2023)Bitterwolf, Mueller, and
  Hein]{bitterwolf2023ninco}
Julian Bitterwolf, Maximilian Mueller, and Matthias Hein.
\newblock In or out? {F}ixing imagenet out-of-distribution detection
  evaluation.
\newblock In \emph{International Conference on Machine Learning (ICML)}, 2023.

\bibitem[Bolya et~al.(2025)Bolya, Huang, Sun, Cho, Madotto,
  et~al.]{bolya2025pe}
Daniel Bolya, Po-Yao Huang, Peize Sun, Jang~Hyun Cho, Andrea Madotto, et~al.
\newblock Perception encoder: The best visual embeddings are not at the output
  of the network.
\newblock \emph{arXiv preprint arXiv:2504.13181}, 2025.

\bibitem[Cao et~al.(2024)Cao, Zhong, Zhou, Liu, Liu, and Han]{cao2024eoe}
Chentao Cao, Zhun Zhong, Zhanke Zhou, Yang Liu, Tongliang Liu, and Bo Han.
\newblock Envisioning outlier exposure by large language models for
  out-of-distribution detection.
\newblock In \emph{International Conference on Machine Learning (ICML)}, 2024.

\bibitem[Cherti et~al.(2023)Cherti, Beaumont, Wightman, Wortsman, Ilharco,
  Gordon, Schuhmann, Schmidt, and Jitsev]{cherti2023openclip}
Mehdi Cherti, Romain Beaumont, Ross Wightman, Mitchell Wortsman, Gabriel
  Ilharco, Cade Gordon, Christoph Schuhmann, Ludwig Schmidt, and Jenia Jitsev.
\newblock Reproducible scaling laws for contrastive language-image learning.
\newblock In \emph{IEEE/CVF Conference on Computer Vision and Pattern
  Recognition (CVPR)}, 2023.

\bibitem[Chuang et~al.(2025)Chuang, Li, Wang, Yeh, Lyu,
  et~al.]{chuang2025metaclip2}
Yung-Sung Chuang, Yang Li, Dong Wang, Ching-Feng Yeh, Kehan Lyu, et~al.
\newblock {MetaCLIP 2}: A worldwide scaling recipe.
\newblock \emph{arXiv preprint arXiv:2507.22062}, 2025.

\bibitem[Cimpoi et~al.(2014)Cimpoi, Maji, Kokkinos, Mohamed, and
  Vedaldi]{cimpoi2014dtd}
Mircea Cimpoi, Subhransu Maji, Iasonas Kokkinos, Sammy Mohamed, and Andrea
  Vedaldi.
\newblock Describing textures in the wild.
\newblock In \emph{IEEE Conference on Computer Vision and Pattern Recognition
  (CVPR)}, 2014.

\bibitem[Deng et~al.(2009)Deng, Dong, Socher, Li, Li, and
  Fei-Fei]{deng2009imagenet}
Jia Deng, Wei Dong, Richard Socher, Li-Jia Li, Kai Li, and Li Fei-Fei.
\newblock {ImageNet}: A large-scale hierarchical image database.
\newblock In \emph{IEEE Conference on Computer Vision and Pattern Recognition
  (CVPR)}, 2009.

\bibitem[Djurisic et~al.(2023)Djurisic, Bozanic, Ashok, and
  Liu]{djurisic2023ash}
Andrija Djurisic, Nebojsa Bozanic, Arjun Ashok, and Rosanne Liu.
\newblock Extremely simple activation shaping for out-of-distribution
  detection.
\newblock In \emph{International Conference on Learning Representations
  (ICLR)}, 2023.

\bibitem[Dosovitskiy et~al.(2021)Dosovitskiy, Beyer, Kolesnikov, Weissenborn,
  Zhai, Unterthiner, Dehghani, Minderer, Heigold, Gelly, Uszkoreit, and
  Houlsby]{dosovitskiy2021vit}
Alexey Dosovitskiy, Lucas Beyer, Alexander Kolesnikov, Dirk Weissenborn,
  Xiaohua Zhai, Thomas Unterthiner, Mostafa Dehghani, Matthias Minderer, Georg
  Heigold, Sylvain Gelly, Jakob Uszkoreit, and Neil Houlsby.
\newblock An image is worth 16x16 words: Transformers for image recognition at
  scale.
\newblock In \emph{International Conference on Learning Representations
  (ICLR)}, 2021.

\bibitem[Fang et~al.(2024)Fang, Sun, Wang, Huang, Wang, and Cao]{fang2024eva02}
Yuxin Fang, Quan Sun, Xinggang Wang, Tiejun Huang, Xinlong Wang, and Yue Cao.
\newblock {EVA-02}: A visual representation for neon genesis.
\newblock \emph{Image and Vision Computing}, 2024.

\bibitem[Fini et~al.(2025)Fini, Shukor, Li, Dufter, Klein,
  et~al.]{fini2024aimv2}
Enrico Fini, Mustafa Shukor, Xiujun Li, Philipp Dufter, Michal Klein, et~al.
\newblock Multimodal autoregressive pre-training of large vision encoders.
\newblock In \emph{IEEE/CVF Conference on Computer Vision and Pattern
  Recognition (CVPR)}, 2025.

\bibitem[Galil et~al.(2023)Galil, Dabbah, and El-Yaniv]{galil2023cood}
Ido Galil, Mohammed Dabbah, and Ran El-Yaniv.
\newblock A framework for benchmarking class-out-of-distribution detection and
  its application to {ImageNet}.
\newblock In \emph{International Conference on Learning Representations
  (ICLR)}, 2023.

\bibitem[He et~al.(2016)He, Zhang, Ren, and Sun]{he2016resnet}
Kaiming He, Xiangyu Zhang, Shaoqing Ren, and Jian Sun.
\newblock Deep residual learning for image recognition.
\newblock In \emph{IEEE Conference on Computer Vision and Pattern Recognition
  (CVPR)}, 2016.

\bibitem[He et~al.(2022)He, Chen, Xie, Li, Doll{\'a}r, and Girshick]{he2022mae}
Kaiming He, Xinlei Chen, Saining Xie, Yanghao Li, Piotr Doll{\'a}r, and Ross
  Girshick.
\newblock Masked autoencoders are scalable vision learners.
\newblock In \emph{IEEE/CVF Conference on Computer Vision and Pattern
  Recognition (CVPR)}, 2022.

\bibitem[Hendrycks and Gimpel(2017)]{hendrycks2017msp}
Dan Hendrycks and Kevin Gimpel.
\newblock A baseline for detecting misclassified and out-of-distribution
  examples in neural networks.
\newblock In \emph{International Conference on Learning Representations
  (ICLR)}, 2017.

\bibitem[Hendrycks et~al.(2021)Hendrycks, Zhao, Basart, Steinhardt, and
  Song]{hendrycks2021imageneto}
Dan Hendrycks, Kevin Zhao, Steven Basart, Jacob Steinhardt, and Dawn Song.
\newblock Natural adversarial examples.
\newblock In \emph{IEEE/CVF Conference on Computer Vision and Pattern
  Recognition (CVPR)}, 2021.

\bibitem[Hendrycks et~al.(2022)Hendrycks, Basart, Mazeika, Zou, Kwon,
  Mostajabi, Steinhardt, and Song]{hendrycks2022scaling}
Dan Hendrycks, Steven Basart, Mantas Mazeika, Andy Zou, Joe Kwon, Mohammadreza
  Mostajabi, Jacob Steinhardt, and Dawn Song.
\newblock Scaling out-of-distribution detection for real-world settings.
\newblock In \emph{International Conference on Machine Learning (ICML)}, 2022.

\bibitem[Huang and Yixuan(2021)]{huang2021mos}
Rui Huang and Li Yixuan.
\newblock {MOS}: Towards scaling out-of-distribution detection for large
  semantic space.
\newblock In \emph{Proceedings of the IEEE/CVF Conference on Computer Vision
  and Pattern Recognition (CVPR)}, pages 8710--8719, 2021.

\bibitem[Jiang et~al.(2024)Jiang, Liu, Fang, Chen, Liu, Zheng, and
  Han]{jiang2024neglabel}
Xue Jiang, Feng Liu, Zhen Fang, Hong Chen, Tongliang Liu, Feng Zheng, and Bo
  Han.
\newblock Negative label guided {OOD} detection with pretrained vision-language
  models.
\newblock In \emph{International Conference on Learning Representations
  (ICLR)}, 2024.

\bibitem[Kisel et~al.(2026)Kisel, Volkov, Janouskova, and
  Matas]{kisel2026multimodal}
Nikita Kisel, Illia Volkov, Klara Janouskova, and Jiri Matas.
\newblock Multimodal large language models as image classifiers.
\newblock \emph{arXiv preprint arXiv:2603.06578}, 2026.

\bibitem[Krizhevsky(2009)]{krizhevsky2009cifar}
Alex Krizhevsky.
\newblock Learning multiple layers of features from tiny images.
\newblock Technical report, University of Toronto, 2009.

\bibitem[Lee et~al.(2018)Lee, Lee, Lee, and Shin]{lee2018mahalanobis}
Kimin Lee, Kibok Lee, Honglak Lee, and Jinwoo Shin.
\newblock A simple unified framework for detecting out-of-distribution samples
  and adversarial attacks.
\newblock In \emph{Advances in Neural Information Processing Systems
  (NeurIPS)}, 2018.

\bibitem[Liu and Qin(2024)]{liu2024fdbd}
Litian Liu and Yao Qin.
\newblock Fast decision boundary based out-of-distribution detector.
\newblock In \emph{International Conference on Machine Learning (ICML)}, 2024.

\bibitem[Liu et~al.(2020)Liu, Wang, Owens, and Yixuan]{liu2020energy}
Weitang Liu, Xiaoyun Wang, John Owens, and Li Yixuan.
\newblock Energy-based out-of-distribution detection.
\newblock In \emph{Advances in Neural Information Processing Systems
  (NeurIPS)}, 2020.

\bibitem[Liu et~al.(2023)Liu, Lochman, and Zach]{liu2023gen}
Xixi Liu, Yaroslava Lochman, and Christopher Zach.
\newblock {GEN}: Pushing the limits of softmax-based out-of-distribution
  detection.
\newblock In \emph{IEEE/CVF Conference on Computer Vision and Pattern
  Recognition (CVPR)}, 2023.

\bibitem[Liu et~al.(2024)Liu, Tian, Li, Ma, and Wang]{liu2024nac}
Yibing Liu, Chris~Xing Tian, Haoliang Li, Lei Ma, and Shiqi Wang.
\newblock Neuron activation coverage: Rethinking out-of-distribution detection
  and generalization.
\newblock In \emph{International Conference on Learning Representations
  (ICLR)}, 2024.

\bibitem[Liu et~al.(2022{\natexlab{a}})Liu, Hu, Lin, Yao, Xie, Wei, Ning, Cao,
  Zhang, Dong, Wei, and Guo]{liu2022swinv2}
Ze Liu, Han Hu, Yutong Lin, Zhuliang Yao, Zhenda Xie, Yixuan Wei, Jia Ning, Yue
  Cao, Zheng Zhang, Li Dong, Furu Wei, and Baining Guo.
\newblock Swin transformer v2: Scaling up capacity and resolution.
\newblock In \emph{IEEE/CVF Conference on Computer Vision and Pattern
  Recognition (CVPR)}, 2022{\natexlab{a}}.

\bibitem[Liu et~al.(2022{\natexlab{b}})Liu, Mao, Wu, Feichtenhofer, Darrell,
  and Xie]{liu2022convnext}
Zhuang Liu, Hanzi Mao, Chao-Yuan Wu, Christoph Feichtenhofer, Trevor Darrell,
  and Saining Xie.
\newblock A convnet for the 2020s.
\newblock In \emph{IEEE/CVF Conference on Computer Vision and Pattern
  Recognition (CVPR)}, 2022{\natexlab{b}}.

\bibitem[Miller(1995)]{miller1995wordnet}
George~A. Miller.
\newblock {WordNet}: A lexical database for english.
\newblock \emph{Communications of the ACM}, 38\penalty0 (11):\penalty0 39--41,
  1995.

\bibitem[Ming et~al.(2022)Ming, Cai, Gu, Sun, Li, and Li]{ming2022mcm}
Yifei Ming, Ziyang Cai, Jiuxiang Gu, Yiyou Sun, Wei Li, and Yixuan Li.
\newblock Delving into out-of-distribution detection with vision-language
  representations.
\newblock In \emph{Advances in Neural Information Processing Systems
  (NeurIPS)}, 2022.

\bibitem[Miyai et~al.(2025)Miyai, Yu, Irie, and Aizawa]{miyai2025glmcm}
Atsuyuki Miyai, Qing Yu, Go Irie, and Kiyoharu Aizawa.
\newblock {GL-MCM}: Global and local maximum concept matching for zero-shot
  out-of-distribution detection.
\newblock \emph{International Journal of Computer Vision (IJCV)}, 2025.

\bibitem[M{\"u}ller and Hein(2025)]{mueller2025mahalanobispp}
Maximilian M{\"u}ller and Matthias Hein.
\newblock Mahalanobis++: Improving {OOD} detection via feature normalization.
\newblock In \emph{International Conference on Machine Learning (ICML)}, 2025.

\bibitem[Northcutt et~al.(2021)Northcutt, Athalye, and
  Mueller]{northcutt2021labelerrors}
Curtis~G. Northcutt, Anish Athalye, and Jonas Mueller.
\newblock Pervasive label errors in test sets destabilize machine learning
  benchmarks.
\newblock In \emph{Advances in Neural Information Processing Systems (NeurIPS)
  Datasets and Benchmarks Track}, 2021.

\bibitem[Oquab et~al.(2024)Oquab, Darcet, Moutakanni, Vo, Szafraniec, Khalidov,
  Fernandez, Haziza, Massa, El-Nouby, et~al.]{oquab2023dinov2}
Maxime Oquab, Timoth{\'e}e Darcet, Th{\'e}o Moutakanni, Huy~V. Vo, Marc
  Szafraniec, Vasil Khalidov, Pierre Fernandez, Daniel Haziza, Francisco Massa,
  Alaaeldin El-Nouby, et~al.
\newblock {DINOv2}: Learning robust visual features without supervision.
\newblock \emph{Transactions on Machine Learning Research}, 2024.

\bibitem[Park et~al.(2023)Park, Jung, and Teoh]{park2023nnguide}
Jaewoo Park, Yoon~Gyo Jung, and Andrew Beng~Jin Teoh.
\newblock Nearest neighbor guidance for out-of-distribution detection.
\newblock In \emph{IEEE/CVF International Conference on Computer Vision
  (ICCV)}, 2023.

\bibitem[Radford et~al.(2021)Radford, Kim, Hallacy, Ramesh, Goh, Agarwal,
  Sastry, Askell, Mishkin, Clark, Krueger, and Sutskever]{radford2021clip}
Alec Radford, Jong~Wook Kim, Chris Hallacy, Aditya Ramesh, Gabriel Goh,
  Sandhini Agarwal, Girish Sastry, Amanda Askell, Pamela Mishkin, Jack Clark,
  Gretchen Krueger, and Ilya Sutskever.
\newblock Learning transferable visual models from natural language
  supervision.
\newblock In \emph{International Conference on Machine Learning (ICML)}, 2021.

\bibitem[Ren et~al.(2021)Ren, Fort, Liu, Roy, Padhy, and
  Lakshminarayanan]{ren2021rmaha}
Jie Ren, Stanislav Fort, Jeremiah Liu, Abhijit~Guha Roy, Shreyas Padhy, and
  Balaji Lakshminarayanan.
\newblock A simple fix to mahalanobis distance for improving near-{OOD}
  detection.
\newblock \emph{arXiv preprint arXiv:2106.09022}, 2021.

\bibitem[Russakovsky et~al.(2015)Russakovsky, Deng, Su, Krause, Satheesh, Ma,
  Huang, Karpathy, Khosla, Bernstein, Berg, and
  Fei-Fei]{russakovsky2015imagenet}
Olga Russakovsky, Jia Deng, Hao Su, Jonathan Krause, Sanjeev Satheesh, Sean Ma,
  Zhiheng Huang, Andrej Karpathy, Aditya Khosla, Michael Bernstein,
  Alexander~C. Berg, and Li Fei-Fei.
\newblock {ImageNet} large scale visual recognition challenge.
\newblock \emph{International Journal of Computer Vision (IJCV)}, 115\penalty0
  (3):\penalty0 211--252, 2015.

\bibitem[Sim{\'e}oni et~al.(2025)Sim{\'e}oni, Vo, Seitzer, Baldassarre, Oquab,
  et~al.]{simeoni2025dinov3}
Oriane Sim{\'e}oni, Huy~V. Vo, Maximilian Seitzer, Federico Baldassarre, Maxime
  Oquab, et~al.
\newblock {DINOv3}.
\newblock \emph{arXiv preprint arXiv:2508.10104}, 2025.

\bibitem[Sun and Li(2022)]{sun2022dice}
Yiyou Sun and Yixuan Li.
\newblock {DICE}: Leveraging sparsification for out-of-distribution detection.
\newblock In \emph{European Conference on Computer Vision (ECCV)}, 2022.

\bibitem[Sun et~al.(2021)Sun, Guo, and Li]{sun2021react}
Yiyou Sun, Chuan Guo, and Yixuan Li.
\newblock {ReAct}: Out-of-distribution detection with rectified activations.
\newblock In \emph{Advances in Neural Information Processing Systems
  (NeurIPS)}, 2021.

\bibitem[Sun et~al.(2022)Sun, Ming, Zhu, and Li]{sun2022knn}
Yiyou Sun, Yifei Ming, Xiaojin Zhu, and Yixuan Li.
\newblock Out-of-distribution detection with deep nearest neighbors.
\newblock In \emph{International Conference on Machine Learning (ICML)}, 2022.

\bibitem[Tschannen et~al.(2025)Tschannen, Gritsenko, Wang, Naeem,
  Alabdulmohsin, et~al.]{tschannen2025siglip2}
Michael Tschannen, Alexey Gritsenko, Xiao Wang, Muhammad~Ferjad Naeem, Ibrahim
  Alabdulmohsin, et~al.
\newblock {SigLIP 2}: Multilingual vision-language encoders with improved
  semantic understanding, localization, and dense features.
\newblock \emph{arXiv preprint arXiv:2502.14786}, 2025.

\bibitem[Vaze et~al.(2022)Vaze, Han, Vedaldi, and Zisserman]{vaze2022ssb}
Sagar Vaze, Kai Han, Andrea Vedaldi, and Andrew Zisserman.
\newblock Open-set recognition: A good closed-set classifier is all you need?
\newblock In \emph{International Conference on Learning Representations
  (ICLR)}, 2022.

\bibitem[Volkov et~al.(2025)Volkov, Kisel, Janouskova, and
  Matas]{volkov2025vlmembeddings}
Illia Volkov, Nikita Kisel, Klara Janouskova, and Jiri Matas.
\newblock Image recognition with vision and language embeddings of {VLMs}.
\newblock In \emph{36th British Machine Vision Conference 2025, {BMVC} 2025,
  Sheffield, UK, November 24-27, 2025}. BMVA, 2025.

\bibitem[Volkov et~al.(2026)Volkov, Kisel, Mishkina, Janouskova, and
  Matas]{reimagenet}
Illia Volkov, Nikita Kisel, Tetiana Mishkina, Klara Janouskova, and Jiri Matas.
\newblock Doomed to re-annotate, forever: The {ImageNet} story.
\newblock \emph{arXiv preprint arXiv:2608.13783}, 2026.

\bibitem[Wang et~al.(2022{\natexlab{a}})Wang, Li, Feng, and Zhang]{wang2022vim}
Haoqi Wang, Zhizhong Li, Litong Feng, and Wayne Zhang.
\newblock {ViM}: Out-of-distribution with virtual-logit matching.
\newblock In \emph{IEEE/CVF Conference on Computer Vision and Pattern
  Recognition (CVPR)}, 2022{\natexlab{a}}.

\bibitem[Wang et~al.(2022{\natexlab{b}})Wang, Zhang, Zhu, Zheng, Li, Smola, and
  Wang]{wang2022pascl}
Haotao Wang, Aston Zhang, Yi Zhu, Shuai Zheng, Mu Li, Alex Smola, and Zhangyang
  Wang.
\newblock Partial and asymmetric contrastive learning for out-of-distribution
  detection in long-tailed recognition.
\newblock In \emph{International Conference on Machine Learning (ICML)},
  2022{\natexlab{b}}.

\bibitem[Wang et~al.(2023)Wang, Li, Yao, and Li]{wang2023clipn}
Haisong Wang, Yiwei Li, Huadong Yao, and Xiaomeng Li.
\newblock {CLIPN} for zero-shot {OOD} detection: Teaching {CLIP} to say no.
\newblock In \emph{Proceedings of the IEEE/CVF International Conference on
  Computer Vision (ICCV)}, pages 1802--1812, 2023.

\bibitem[Woo et~al.(2023)Woo, Debnath, Hu, Chen, Liu, Kweon, and
  Xie]{woo2023convnextv2}
Sanghyun Woo, Shoubhik Debnath, Ronghang Hu, Xinlei Chen, Zhuang Liu, In~So
  Kweon, and Saining Xie.
\newblock {ConvNeXt V2}: Co-designing and scaling convnets with masked
  autoencoders.
\newblock In \emph{IEEE/CVF Conference on Computer Vision and Pattern
  Recognition (CVPR)}, 2023.

\bibitem[Xu et~al.(2024)Xu, Chen, Franchi, and Yao]{xu2024scale}
Kai Xu, Rongyu Chen, Gianni Franchi, and Angela Yao.
\newblock Scaling for training time and post-hoc out-of-distribution detection
  enhancement.
\newblock In \emph{International Conference on Learning Representations
  (ICLR)}, 2024.

\bibitem[Yang et~al.(2022)Yang, Wang, Zou, Zhou, et~al.]{yang2022openood}
Jingkang Yang, Pengyun Wang, Dejian Zou, Zitang Zhou, et~al.
\newblock {OpenOOD}: Benchmarking generalized out-of-distribution detection.
\newblock In \emph{Advances in Neural Information Processing Systems
  (NeurIPS)}, 2022.

\bibitem[Yang et~al.(2023)Yang, Zhang, and Russakovsky]{imagenetood}
William Yang, Byron Zhang, and Olga Russakovsky.
\newblock {ImageNet-OOD}: Deciphering modern out-of-distribution detection
  algorithms.
\newblock \emph{arXiv preprint arXiv:2310.01755}, 2023.

\bibitem[Zhai et~al.(2023)Zhai, Mustafa, Kolesnikov, and Beyer]{zhai2023siglip}
Xiaohua Zhai, Basil Mustafa, Alexander Kolesnikov, and Lucas Beyer.
\newblock Sigmoid loss for language image pre-training.
\newblock In \emph{IEEE/CVF International Conference on Computer Vision
  (ICCV)}, 2023.

\bibitem[Zhang et~al.(2023)Zhang, Yang, Wang, et~al.]{zhang2023openoodv15}
Jingyang Zhang, Jingkang Yang, Pengyun Wang, et~al.
\newblock {OpenOOD} v1.5: Enhanced benchmark for out-of-distribution detection.
\newblock \emph{arXiv preprint arXiv:2306.09301}, 2023.

\end{thebibliography}
}

\appendix
\maketitlesupplementary

\renewcommand{\thetable}{A\arabic{table}}
\renewcommand{\thefigure}{A\arabic{figure}}
\setcounter{table}{0}
\setcounter{figure}{0}

\setcounter{topnumber}{3}
\setcounter{bottomnumber}{2}
\setcounter{totalnumber}{5}
\setcounter{dbltopnumber}{3}
\renewcommand{\topfraction}{0.92}
\renewcommand{\bottomfraction}{0.85}
\renewcommand{\dbltopfraction}{0.92}
\renewcommand{\textfraction}{0.08}
\renewcommand{\floatpagefraction}{0.50}
\renewcommand{\dblfloatpagefraction}{0.50}

Throughout the supplementary material, we use the same ID set, OOD set, and ID-anchored FPR@95 threshold as in the main paper.
\Cref{supp:results} provides the complete evaluation results,
\cref{supp:vocab} analyzes the negative-vocabulary design,
\cref{supp:size} studies the effect of OOD sample size,
\cref{supp:gallery} shows more examples from our dataset,
\cref{supp:model-zoo} extends our study of supervised models,
and \cref{supp:impl,supp:prompts} document implementation details, model checkpoints, and prompts.

\section{Full Evaluation Results}
\label{supp:results}

\providecommand{\g}[1]{{\scriptsize\color{gray}#1}}
\begin{table*}[b!]
\centering
\small
\setlength{\tabcolsep}{0pt}
\renewcommand{\arraystretch}{1.15}
\caption{
    \textbf{All 21 post-hoc detectors on \bench{}.}
    Performance of every supervised classifier of \cref{tab:supervised} with every detector, as \mbox{FPR@95 (\%, $\downarrow$)\,\g{/\,AUROC (\%, $\uparrow$)}}. Per model, the best FPR@95 is in \textbf{bold} and the best AUROC \underline{underlined}; \cref{tab:supervised} reports only the best per column (model). Groups: logit-based, activation- and weight-based, feature-based, hybrid. $\dagger$ marks a linear probe on a frozen backbone.
}
\label{tab:supp-all-detectors}
\begin{tabular*}{\textwidth}{@{}l@{\extracolsep{\fill}}cccccccc@{}}
\toprule
Detector & \multicolumn{1}{c}{ResNet50} & \multicolumn{1}{c}{SwinV2-B-256} & \multicolumn{1}{c}{ConvNeXt-B} & \multicolumn{1}{c}{ConvNeXtV2-L} & \multicolumn{1}{c}{ViT-B/16-384} & \multicolumn{1}{c}{DINOv2-g/14 $\dagger$} & \multicolumn{1}{c}{EVA02-L-448} & \multicolumn{1}{c}{DINOv3-7B $\dagger$} \\
\midrule
MSP & 85.8\,\g{/\,71.8} & 78.0\,\g{/\,71.7} & 78.0\,\g{/\,73.7} & 77.3\,\g{/\,73.5} & 78.3\,\g{/\,75.5} & 79.1\,\g{/\,75.7} & 70.5\,\g{/\,74.4} & 80.2\,\g{/\,75.3} \\
MaxLogit & 87.0\,\g{/\,71.8} & 73.6\,\g{/\,70.8} & 73.1\,\g{/\,72.8} & 71.5\,\g{/\,71.5} & 74.4\,\g{/\,78.6} & 72.1\,\g{/\,77.3} & 67.8\,\g{/\,72.0} & 71.6\,\g{/\,79.4} \\
KL-Match. & 84.6\,\g{/\,70.0} & 82.0\,\g{/\,72.4} & 82.0\,\g{/\,74.3} & 78.6\,\g{/\,76.3} & 80.0\,\g{/\,73.8} & 78.2\,\g{/\,75.1} & 75.4\,\g{/\,80.1} & 78.2\,\g{/\,75.2} \\
Energy & 88.9\,\g{/\,71.5} & 73.4\,\g{/\,69.1} & 70.8\,\g{/\,71.6} & 69.9\,\g{/\,70.8} & 74.5\,\g{/\,78.3} & 71.1\,\g{/\,77.0} & 67.2\,\g{/\,68.9} & 70.8\,\g{/\,79.2} \\
GEN & 86.3\,\g{/\,71.4} & 77.3\,\g{/\,72.8} & 74.2\,\g{/\,75.5} & 71.0\,\g{/\,75.1} & 76.0\,\g{/\,79.2} & 73.9\,\g{/\,\underline{77.5}} & 67.0\,\g{/\,77.9} & 73.1\,\g{/\,79.6} \\
\midrule
E+ReAct & 96.9\,\g{/\,54.5} & 75.7\,\g{/\,70.7} & 71.1\,\g{/\,72.8} & 70.5\,\g{/\,72.1} & 78.5\,\g{/\,76.2} & 71.1\,\g{/\,77.1} & 66.7\,\g{/\,70.4} & 71.1\,\g{/\,79.3} \\
ASH-S & 86.4\,\g{/\,70.4} & 95.1\,\g{/\,62.3} & 95.9\,\g{/\,55.7} & 86.7\,\g{/\,66.9} & 95.7\,\g{/\,56.0} & 96.3\,\g{/\,47.6} & 87.2\,\g{/\,65.4} & 94.2\,\g{/\,52.1} \\
SCALE & 86.3\,\g{/\,71.9} & 93.7\,\g{/\,46.7} & 77.7\,\g{/\,72.0} & 73.9\,\g{/\,75.0} & 70.8\,\g{/\,80.4} & 71.6\,\g{/\,77.0} & 68.2\,\g{/\,63.9} & 72.5\,\g{/\,78.3} \\
DICE & 88.4\,\g{/\,66.0} & 88.4\,\g{/\,51.7} & 94.4\,\g{/\,45.6} & 86.1\,\g{/\,65.3} & 97.3\,\g{/\,47.3} & 95.4\,\g{/\,45.3} & 84.7\,\g{/\,58.2} & 88.1\,\g{/\,64.8} \\
\midrule
Maha & 91.0\,\g{/\,61.5} & 84.6\,\g{/\,74.1} & 79.7\,\g{/\,76.4} & 75.6\,\g{/\,81.2} & 69.2\,\g{/\,\underline{81.4}} & 69.9\,\g{/\,74.4} & 66.7\,\g{/\,83.3} & 93.0\,\g{/\,70.8} \\
rMaha & 83.7\,\g{/\,73.0} & 81.7\,\g{/\,76.0} & 76.3\,\g{/\,79.0} & 67.9\,\g{/\,82.1} & 70.2\,\g{/\,81.2} & 69.3\,\g{/\,75.4} & 66.4\,\g{/\,83.9} & 70.2\,\g{/\,80.0} \\
Maha++ & 82.3\,\g{/\,70.4} & 76.5\,\g{/\,77.0} & 70.7\,\g{/\,79.9} & 68.2\,\g{/\,\underline{83.1}} & \textbf{69.0}\,\g{/\,80.3} & 70.1\,\g{/\,74.1} & \textbf{65.0}\,\g{/\,83.8} & 74.8\,\g{/\,76.3} \\
rMaha++ & \textbf{80.2}\,\g{/\,\underline{74.2}} & 75.6\,\g{/\,\underline{77.9}} & \textbf{69.2}\,\g{/\,\underline{80.3}} & \textbf{67.8}\,\g{/\,82.7} & 70.5\,\g{/\,81.1} & 69.5\,\g{/\,75.4} & \textbf{65.0}\,\g{/\,\underline{84.2}} & \textbf{64.4}\,\g{/\,80.6} \\
$k$-NN & 91.5\,\g{/\,54.1} & 85.5\,\g{/\,66.3} & 81.4\,\g{/\,71.6} & 78.6\,\g{/\,76.9} & 83.5\,\g{/\,69.5} & 76.0\,\g{/\,67.2} & 75.3\,\g{/\,78.3} & 84.0\,\g{/\,73.7} \\
cosine & 86.1\,\g{/\,66.5} & 80.8\,\g{/\,74.4} & 76.8\,\g{/\,78.0} & 71.1\,\g{/\,81.8} & 78.5\,\g{/\,77.8} & 70.7\,\g{/\,73.3} & 67.9\,\g{/\,83.3} & 67.5\,\g{/\,80.8} \\
MCM & 83.1\,\g{/\,71.3} & 80.8\,\g{/\,75.4} & 76.6\,\g{/\,78.2} & 70.7\,\g{/\,81.8} & 77.7\,\g{/\,78.1} & 70.4\,\g{/\,73.4} & 67.9\,\g{/\,83.2} & 67.2\,\g{/\,\underline{81.0}} \\
fDBD & 89.3\,\g{/\,66.6} & 82.6\,\g{/\,70.0} & 78.9\,\g{/\,72.6} & 79.5\,\g{/\,75.8} & 78.8\,\g{/\,78.2} & 73.6\,\g{/\,77.1} & 71.6\,\g{/\,80.4} & 82.1\,\g{/\,74.5} \\
NECO & 87.5\,\g{/\,71.8} & 81.1\,\g{/\,63.0} & 77.6\,\g{/\,67.0} & 77.3\,\g{/\,68.3} & 74.4\,\g{/\,75.8} & 71.8\,\g{/\,71.7} & 71.8\,\g{/\,65.7} & 78.6\,\g{/\,76.6} \\
NAC & 96.6\,\g{/\,55.0} & \textbf{73.3}\,\g{/\,67.5} & 72.4\,\g{/\,71.4} & 69.2\,\g{/\,68.0} & 73.0\,\g{/\,79.2} & 74.8\,\g{/\,71.0} & 66.9\,\g{/\,70.0} & 78.2\,\g{/\,68.6} \\
\midrule
ViM & 87.0\,\g{/\,71.4} & 82.7\,\g{/\,74.1} & 77.7\,\g{/\,76.8} & 77.6\,\g{/\,80.2} & 73.9\,\g{/\,79.5} & \textbf{68.7}\,\g{/\,77.4} & 68.2\,\g{/\,81.9} & 79.8\,\g{/\,79.4} \\
NNGuide & 88.9\,\g{/\,71.6} & 81.8\,\g{/\,73.2} & 75.7\,\g{/\,78.1} & 72.7\,\g{/\,80.2} & 79.4\,\g{/\,75.2} & 73.6\,\g{/\,74.2} & 72.4\,\g{/\,81.1} & 75.4\,\g{/\,78.8} \\
\bottomrule
\end{tabular*}\end{table*}

\Cref{tab:supp-all-detectors} expands \cref{tab:supervised} to all 21 post-hoc detectors.
A Mahalanobis-family score is the best detector or within 2.3 points of it for every backbone.
The activation- and weight-reshaping scores are the weakest group.
The lowest observed FPR@95 is 64.4\%. No model--detector configuration attains a lower value, indicating that the benchmark remains difficult across all evaluated detector--backbone pairs.

\Cref{tab:supp-per-model-datasets} expands \cref{tab:eva-siglip-subsets} to all eight supervised models of \cref{tab:supervised}: the six fine-tuned classifiers and the two linear probes on frozen backbones.
In terms of FPR@95, \bench{} is the most difficult evaluated dataset for every model in \cref{tab:supp-per-model-datasets}, with margins of 7.1 (ResNet50) to 18.2 points (SwinV2-B-256).
EVA02-L-448, the model that performs best on other datasets, achieves mean FPR@95 of 21.2\% on other datasets, but 65.0\% on ours.

\Cref{tab:supp-full-vs-clean} compares popular OOD datasets in full size (\cref{tab:eva-siglip-subsets}) with their ``clean'' sub-samples, as constructed by NINCO \cite{bitterwolf2023ninco}.
Cleaning lowers FPR@95 on all 11 datasets (22.6\% to 12.2\% on average), varying from 0.7 points down on iNat-OOD Plants to 25.1 on Species and 22.7 on SSB easy (0.3 percent subset).

\providecommand{\g}[1]{{\scriptsize\color{gray}#1}}
\begin{table*}[t]\centering\footnotesize\setlength{\tabcolsep}{0pt}\renewcommand{\arraystretch}{1.05}
\caption{\textbf{Every supervised classifier on every OOD dataset.}
Best \mbox{FPR@95 (\%, $\downarrow$)\,\g{/\,AUROC (\%, $\uparrow$)}} of the 21 post-hoc detectors, taken per model and dataset, the two metrics selected independently.
\Cref{tab:eva-siglip-subsets} reports only on EVA02-L-448 (see also \cref{fig:fpr_ours_vs_popular}).
$\dagger$ marks a linear probe on a frozen backbone.}
\label{tab:supp-per-model-datasets}
\begin{tabular*}{\textwidth}{@{}l@{\extracolsep{\fill}}r@{}lr@{}lr@{}lr@{}lr@{}lr@{}lr@{}lr@{}lr@{}l@{}}\toprule
Dataset & \multicolumn{2}{c}{ResNet50} & \multicolumn{2}{c}{SwinV2-B-256} & \multicolumn{2}{c}{ConvNeXt-B} & \multicolumn{2}{c}{ConvNeXtV2-L} & \multicolumn{2}{c}{ViT-B/16-384} & \multicolumn{2}{c}{DINOv2-g/14 $\dagger$} & \multicolumn{2}{c}{EVA02-L-448} & \multicolumn{2}{c}{DINOv3-7B $\dagger$} & \multicolumn{2}{c}{Mean} \\ \midrule
iNat-OOD Plants & 31.4 & \g{/\,93.8} & 1.3 & \g{/\,99.6} & 1.7 & \g{/\,99.5} & 0.8 & \g{/\,99.8} & 1.5 & \g{/\,99.7} & 0.9 & \g{/\,99.6} & 1.2 & \g{/\,99.8} & 1.4 & \g{/\,99.6} & 5.0 & \g{/\,98.9} \\
OpenImage-O & 44.9 & \g{/\,91.9} & 9.7 & \g{/\,98.0} & 8.5 & \g{/\,98.2} & 5.5 & \g{/\,98.8} & 9.8 & \g{/\,98.1} & 10.0 & \g{/\,97.8} & 5.8 & \g{/\,98.8} & 8.6 & \g{/\,97.9} & 12.8 & \g{/\,97.4} \\
360OpenSet & 55.0 & \g{/\,86.9} & 19.5 & \g{/\,95.5} & 17.4 & \g{/\,95.8} & 12.8 & \g{/\,96.9} & 15.8 & \g{/\,96.5} & 15.6 & \g{/\,95.9} & 11.9 & \g{/\,97.0} & 16.5 & \g{/\,95.5} & 20.6 & \g{/\,95.0} \\
COOD & 56.8 & \g{/\,86.4} & 24.3 & \g{/\,93.6} & 20.8 & \g{/\,94.9} & 15.3 & \g{/\,96.4} & 20.9 & \g{/\,95.2} & 21.3 & \g{/\,94.3} & 13.8 & \g{/\,96.6} & 22.2 & \g{/\,94.3} & 24.4 & \g{/\,94.0} \\
ImageNet-O & 69.2 & \g{/\,79.5} & 38.4 & \g{/\,91.5} & 32.8 & \g{/\,93.2} & 22.7 & \g{/\,94.7} & 31.9 & \g{/\,93.6} & 30.6 & \g{/\,93.4} & 18.0 & \g{/\,95.8} & 30.6 & \g{/\,92.6} & 34.2 & \g{/\,91.8} \\
Textures & 33.7 & \g{/\,91.2} & 28.0 & \g{/\,92.6} & 24.1 & \g{/\,93.9} & 22.0 & \g{/\,94.0} & 22.3 & \g{/\,94.0} & 25.4 & \g{/\,94.1} & 21.2 & \g{/\,94.1} & 22.4 & \g{/\,94.5} & 24.9 & \g{/\,93.5} \\
Places & 64.9 & \g{/\,84.9} & 35.2 & \g{/\,89.9} & 41.1 & \g{/\,90.6} & 30.4 & \g{/\,94.2} & 41.8 & \g{/\,91.0} & 27.1 & \g{/\,93.2} & 25.3 & \g{/\,94.6} & 17.8 & \g{/\,95.7} & 35.5 & \g{/\,91.8} \\
SSB easy & 67.6 & \g{/\,81.5} & 45.5 & \g{/\,87.5} & 43.6 & \g{/\,89.1} & 34.7 & \g{/\,90.9} & 40.6 & \g{/\,89.6} & 46.1 & \g{/\,88.4} & 30.0 & \g{/\,92.1} & 41.8 & \g{/\,88.6} & 43.7 & \g{/\,88.5} \\
SSB hard & 73.1 & \g{/\,78.3} & 47.6 & \g{/\,85.9} & 46.9 & \g{/\,86.8} & 41.8 & \g{/\,88.8} & 49.9 & \g{/\,87.8} & 48.5 & \g{/\,83.7} & 33.1 & \g{/\,92.7} & 51.4 & \g{/\,85.5} & 49.0 & \g{/\,86.2} \\
Species & 63.7 & \g{/\,79.9} & 50.2 & \g{/\,83.3} & 46.5 & \g{/\,84.7} & 36.4 & \g{/\,88.5} & 41.6 & \g{/\,87.9} & 50.0 & \g{/\,80.4} & 40.2 & \g{/\,88.4} & 49.0 & \g{/\,87.4} & 47.2 & \g{/\,85.1} \\
IN1k-OOD & 63.7 & \g{/\,75.7} & 55.0 & \g{/\,81.6} & 52.2 & \g{/\,83.8} & 51.6 & \g{/\,83.5} & 60.8 & \g{/\,82.2} & 60.5 & \g{/\,81.6} & 48.2 & \g{/\,83.3} & 53.0 & \g{/\,81.5} & 55.6 & \g{/\,81.6} \\
\midrule
CLEAN-OOD & 51.6 & \g{/\,87.6} & 23.3 & \g{/\,94.4} & 18.7 & \g{/\,95.6} & 12.0 & \g{/\,97.4} & 18.9 & \g{/\,96.0} & 18.1 & \g{/\,94.7} & 12.2 & \g{/\,97.4} & 17.8 & \g{/\,95.5} & 21.6 & \g{/\,94.8} \\
NINCO & 59.4 & \g{/\,86.5} & 32.0 & \g{/\,92.8} & 26.3 & \g{/\,94.7} & 17.0 & \g{/\,96.6} & 24.2 & \g{/\,95.6} & 29.3 & \g{/\,92.5} & 16.9 & \g{/\,96.7} & 18.1 & \g{/\,96.0} & 27.9 & \g{/\,93.9} \\
ImageNet-OOD & 64.1 & \g{/\,85.6} & 36.5 & \g{/\,91.3} & 35.5 & \g{/\,93.2} & 25.7 & \g{/\,94.5} & 33.5 & \g{/\,93.3} & 40.7 & \g{/\,91.5} & 18.8 & \g{/\,96.2} & 38.4 & \g{/\,90.0} & 36.6 & \g{/\,91.9} \\
\midrule
\textit{Mean} & \textit{57.1} & \g{/\,85.0} & \textit{31.9} & \g{/\,91.3} & \textit{29.7} & \g{/\,92.4} & \textit{23.5} & \g{/\,93.9} & \textit{29.5} & \g{/\,92.9} & \textit{30.3} & \g{/\,91.5} & \textit{21.2} & \g{/\,94.5} & \textit{27.8} & \g{/\,92.5} & \textit{31.4} & \g{/\,91.7} \\
\midrule
\textbf{\bench{}} & \textbf{80.2} & \g{/\,74.2} & \textbf{73.3} & \g{/\,77.9} & \textbf{69.2} & \g{/\,80.3} & \textbf{67.8} & \g{/\,83.1} & \textbf{69.0} & \g{/\,81.4} & \textbf{68.7} & \g{/\,77.5} & \textbf{65.0} & \g{/\,84.2} & \textbf{64.4} & \g{/\,81.0} & \textbf{69.7} & \g{/\,79.9} \\
\bottomrule\end{tabular*}\end{table*}

\begin{table*}[tbp]
\centering
\small
\setlength{\tabcolsep}{0pt}
\renewcommand{\arraystretch}{1.0}
\caption{
    \textbf{The eleven popular OOD datasets in full size and as cleaned sub-samples.}
    Extension of \cref{tab:eva-siglip-subsets} to all eleven datasets.
    Each line pairs the \textbf{full} dataset with the manually cleaned \textbf{sub-sample} of the same dataset released by NINCO~\cite{bitterwolf2023ninco}, which draws 400 images per dataset and removes every image containing an \imnet[1] object; $n$ is the number of OOD images in each.
    Values are the best FPR@95 (\%, $\downarrow$) and AUROC (\%, $\uparrow$) of EVA02-L-448 of its 21 post-hoc detectors, selected independently per dataset and metric, with the best-performing detector named in gray.
    The \% column is the sub-sample as a fraction of the full dataset (in percents).
    The $\Delta$ is sub-sample minus full for each metric.
}
\label{tab:supp-full-vs-clean}
\begin{tabular*}{\textwidth}{
    @{}l@{\extracolsep{\fill}}r r@{\extracolsep{0pt}\hspace{2pt}} l@{\extracolsep{\fill}} r@{\extracolsep{0pt}\hspace{2pt}} l@{\extracolsep{\fill}}r r r@{\extracolsep{0pt}\hspace{2pt}} l@{\extracolsep{\fill}} r@{\extracolsep{0pt}\hspace{2pt}} l@{\extracolsep{\fill}}r r@{}
}
\toprule
& \multicolumn{5}{c}{\textbf{Full} dataset} & \multicolumn{6}{c}{clean \textbf{sub-sample} by NINCO} & \multicolumn{2}{c@{}}{$\Delta$} \\
\cmidrule{2-6}\cmidrule{7-12}\cmidrule{13-14}
Dataset & $n$ & \multicolumn{2}{c}{FPR@95 $\downarrow$} & \multicolumn{2}{c}{AUROC $\uparrow$} & $n$ & \% & \multicolumn{2}{c}{FPR@95 $\downarrow$} & \multicolumn{2}{c}{AUROC $\uparrow$} & FPR & AUROC \\
\midrule
\rule{0pt}{2.4ex}iNat-OOD Plants & 10\,000 & 1.2 & \g{ViM} & 99.8 & \g{ViM} & 383 & 3.8 & 0.5 & \g{ViM} & 99.8 & \g{ViM} & \textminus0.7 & +0.0 \\
OpenImage-O & 17\,632 & 5.8 & \g{Maha++} & 98.8 & \g{Maha++} & 368 & 2.1 & 4.3 & \g{NECO} & 99.0 & \g{Maha++} & \textminus1.5 & +0.2 \\
360OpenSet & 11\,800 & 11.9 & \g{E+ReAct} & 97.0 & \g{Maha++} & 236 & 2.0 & 7.2 & \g{NECO} & 97.9 & \g{Maha++} & \textminus4.7 & +0.9 \\
COOD & 10\,350 & 13.8 & \g{Maha++} & 96.6 & \g{Maha++} & 212 & 2.0 & 9.0 & \g{GEN} & 98.0 & \g{Maha++} & \textminus4.8 & +1.4 \\
ImageNet-O & 2\,000 & 18.0 & \g{Maha++} & 95.8 & \g{Maha++} & 309 & 15.4 & 12.3 & \g{Maha++} & 96.8 & \g{NECO} & \textminus5.7 & +1.0 \\
Textures & 5\,640 & 21.2 & \g{Maha++} & 94.1 & \g{Maha++} & 288 & 5.1 & 9.7 & \g{Maha++} & 98.3 & \g{Maha++} & \textminus11.5 & +4.2 \\
Places & 10\,000 & 25.3 & \g{NNGuide} & 94.6 & \g{ViM} & 153 & 1.5 & 9.2 & \g{Maha++} & 98.3 & \g{ViM} & \textminus16.1 & +3.7 \\
SSB easy & 50\,000 & 30.0 & \g{NECO} & 92.1 & \g{Maha++} & 151 & 0.3 & 7.3 & \g{GEN} & 98.6 & \g{NECO} & \textminus22.7 & +6.5 \\
SSB hard & 49\,000 & 33.1 & \g{ViM} & 92.7 & \g{ViM} & 208 & 0.4 & 24.0 & \g{Energy} & 94.4 & \g{Maha++} & \textminus9.1 & +1.7 \\
Species & 10\,000 & 40.2 & \g{Maha} & 88.4 & \g{ViM} & 172 & 1.7 & 15.1 & \g{Maha++} & 97.4 & \g{ViM} & \textminus25.1 & +9.0 \\
IN1k-OOD & 50\,000 & 48.2 & \g{NAC} & \textbf{83.3} & \g{GEN} & 235 & 0.5 & 35.7 & \g{NAC} & 91.3 & \g{Maha++} & \textminus12.5 & +8.0 \\
\midrule
\bench{} & \nbench{} & \textbf{65.0} & \g{Maha++} & 84.2 & \g{rMaha++} & \nbench{} & 100.0 & \textbf{65.0} & \g{Maha++} & \textbf{84.2} & \g{rMaha++} & — & — \\
\bottomrule
\end{tabular*}
\end{table*}

\twocolumn[{%
    \centering
    \begin{minipage}{0.48\textwidth}
        \centering
        \includegraphics[width=\linewidth]{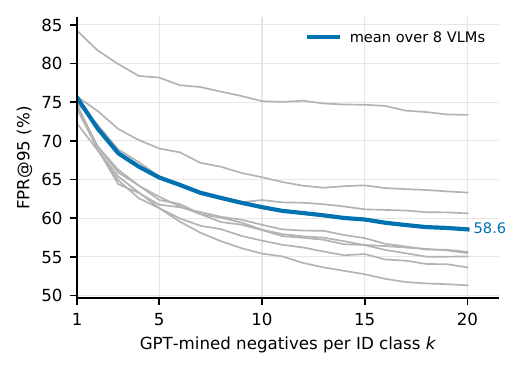}
        \captionof{figure}{\textbf{GPT-generated per ID class siblings.} FPR@95 (\%, $\downarrow$) on \bench{} of NegLabel with GPT-mined per-class siblings as a function of the number $k$ of negatives kept per ID class (gray: the eight VLMs; blue: mean).}
        \label{fig:supp-budget}
    \end{minipage}
    \hfill
    \begin{minipage}{0.48\textwidth}
        \centering
        \includegraphics[width=\linewidth]{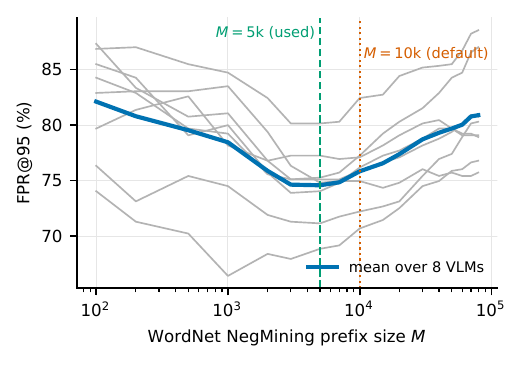}
        \captionof{figure}{\textbf{WordNet negative pool size.} FPR@95 (\%, $\downarrow$) on \bench{} of NegLabel with the NegMined WordNet lexicon as a function of the pool size $M$ (farthest labels first, log scale; gray: the eight VLMs; blue: mean). The default $M{=}10{,}000$ by NegLabel lies past the optimum for every model; \cref{tab:neg-sources} uses $M{=}5000$, the mean of the per-model optima.}
        \label{fig:supp-wordnet}
    \end{minipage}
    \vspace{2mm}
}]

\section{Ablation Study: Negative Vocabulary}
\label{supp:vocab}

The following experiments produce analyses of negative vocabulary construction.

\begin{figure}[t]\centering
\includegraphics[width=.95\columnwidth]{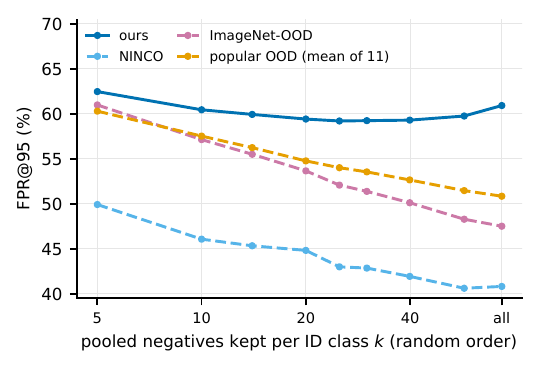}
\caption{
    \textbf{Size of the merged per-ID class negative pool.}
    NegLabel FPR@95 (\%, $\downarrow$; mean over the eight VLMs) with $k$ negatives kept per ID class from the merged
    negative pool of all ten GPT-5.4 generations; ``all'' keeps every candidate (73.6 per ID class
    on average). More negatives keep helping on popular OOD datasets (dashed) but not on
    \bench{} (solid).
}
\label{fig:supp-pooled}
\end{figure}

\noindent\textbf{GPT-generated negative siblings per class (\cref{fig:supp-budget}).}
FPR@95 decreases monotonically with the growing number of negative labels (``siblings'') requested from GPT-5.4 for every VLM.
Each GPT-5.4 generation of negative siblings yields 20 of them per ID class.
The FPR@95 standard deviation between runs requesting 20 negative labels is 1.0--1.5 points.
Merging all ten generations yields 26--138 unique candidates per ID class (73.6 on average). 
We keep $k$ of them, chosen in random order so that $k$ controls only the negative vocabulary size, not a quality ranking in which GPT generates them.
On other OOD datasets, more negatives help monotonically (NINCO:
49.9\% at $k{=}5$, 40.6\% at $k{=}60$).
On \bench{} FPR@95 flattens near 59\% for
$k{=}20$--$40$ (\cref{fig:supp-pooled}) and rises to 60.9\% with all candidates. 
Therefore, the NegLabel GPT experiment from \cref{tab:neg-labels} uses 20 negative siblings per ID class.

\noindent\textbf{WordNet negative pool size (\cref{fig:supp-wordnet}).} \\
For \bench{}, performance is best for WordNet prefixes containing approximately 1,000--15,000 labels, depending on the VLM.
Seven of the eight model-specific optima occur at or below 5,000 labels.
\Cref{tab:neg-sources} uses the fixed value $M{=}5{,}000$, whereas the original NegLabel configuration uses $M{=}10{,}000$.

\noindent\textbf{Combining the two vocabularies (\cref{tab:neg-sources,tab:supp-popular-vocab}).}
Only the WordNet prefix varies here, in whether it is added to the per-class siblings and at which size $M$; the score, ID names, templates and the budget of 20 siblings per class are fixed.
On \bench{} the union gives 58.8\% against 58.6\% for the siblings alone, and $M{=}20{,}000$ raises it to 62.8\%; on NINCO it lowers FPR@95 from 50.4\% to 40.7\% and on ImageNet-OOD from 55.5\% to 53.6\%.
WordNet therefore helps separate OOD images far from \imnet[1], not near-OOD ones of \bench{}.

\noindent\textbf{Negative-label vocabularies on other OOD datasets (\cref{tab:supp-popular-vocab}).}
\Cref{tab:supp-popular-vocab} evaluates the vocabularies of \cref{tab:neg-sources} on the eleven popular OOD datasets, NINCO and ImageNet-OOD.
Which mined vocabulary is best depends on how far the dataset lies from \imnet[1] content: WordNet wins on five sets unrelated to \imnet[1] (\eg, iNat-OOD Plants, Textures, Places), the union wins on the remaining eight, including every \imnet[1]-derived set (SSB, IN1k-OOD, ImageNet-O, ImageNet-OOD) and NINCO, and the per-class siblings alone win only on \bench{}, whose OOD images lie closest to the ID classes.
In the mean over the thirteen OOD datasets, the union is best, 46.8\% against 47.2\% for WordNet and 58.3\% for the siblings.
Nevertheless, the oracle bounds all of them at 4.9--37.4\%, indicating that detection is limited by vocabulary selection rather than by the scoring function.

\noindent\textbf{EOE prompt style (\cref{tab:supp-eoe-prompts}).}
The three EOE prompts differ by about 13 FPR@95 points on \bench{}.
Only the fine-grained prompt improves substantially on MCM, the vocabulary-free baseline (79.7\%), reaching 70.2\% at 500 negative classes (the far-OOD prompt gains 1.3 points, the near-OOD prompt loses 3.7), and it is also best on NINCO and CLEAN-OOD.
On the contrary, on Places it is 12.5 points worse than the near-OOD prompt and on ImageNet-OOD 3.0 worse.
The prompt thus controls which kind of OOD the vocabulary covers.
While the main-paper result uses the fine-grained prompt, which performs best on \bench{}, \cref{tab:supp-eoe-prompts} reports on all three prompts.
EOE is also the least stable vocabulary: its standard deviation across generations at $500$ negatives is 7.8 points.

\begin{table}[t]
\centering
\footnotesize
\setlength{\tabcolsep}{0pt}
\renewcommand{\arraystretch}{1.0}
\caption{
    \textbf{EOE's three released prompts across OOD datasets.}
    FPR@95 (\%, $\downarrow$) of EOE with each of the three prompts released by \cite{cao2024eoe}, at 500 generated categories: mean over the eight VLMs and 30 independent GPT-5.4 generations, $\pm$ the standard deviation over the generations of the mean.
    Best prompt per row in \textbf{bold}, or underlined when it does not beat MCM. \\
    No prompt dominates everywhere: the fine-grained prompt that the main paper uses wins on \bench{} and on the near-OOD sets, and loses to the other two on scene- and texture-like data.
}
\label{tab:supp-eoe-prompts}
\begin{tabular*}{\columnwidth}{
    @{}l@{\extracolsep{\fill}} r@{\hspace{7pt}} r@{\hspace{2pt}}l r@{\hspace{2pt}}l r@{\hspace{2pt}}l@{}
}
\toprule
Dataset & MCM & \multicolumn{2}{c}{far-OOD} & \multicolumn{2}{c}{near-OOD} & \multicolumn{2}{c}{fine-grained} \\
\midrule
\rule{0pt}{2.4ex}\textbf{\bench{}} & 79.7 & 78.4 & {\scriptsize$\pm$6.7} & 83.4 & {\scriptsize$\pm$2.9} & \textbf{70.2} & {\scriptsize$\pm$7.8} \\
\quad fine-grained split & 83.1 & 85.2 & {\scriptsize$\pm$13.1} & 94.9 & {\scriptsize$\pm$4.8} & \textbf{71.4} & {\scriptsize$\pm$15.8} \\
\quad ordinary split & 75.6 & 70.1 & {\scriptsize$\pm$2.5} & 69.4 & {\scriptsize$\pm$3.1} & \textbf{68.8} & {\scriptsize$\pm$3.8} \\
\midrule
NINCO & 60.9 & 55.2 & {\scriptsize$\pm$11.0} & 61.7 & {\scriptsize$\pm$6.3} & \textbf{46.8} & {\scriptsize$\pm$11.1} \\
ImageNet-OOD & 65.7 & 59.6 & {\scriptsize$\pm$4.5} & \textbf{57.0} & {\scriptsize$\pm$3.8} & 60.0 & {\scriptsize$\pm$6.7} \\
CLEAN-OOD & 53.1 & 47.7 & {\scriptsize$\pm$5.8} & 48.7 & {\scriptsize$\pm$5.0} & \textbf{43.6} & {\scriptsize$\pm$8.7} \\
\midrule
iNat-OOD Plants & 41.8 & 33.9 & {\scriptsize$\pm$11.2} & 30.8 & {\scriptsize$\pm$9.9} & \textbf{29.8} & {\scriptsize$\pm$15.2} \\
OpenImage-O & 44.9 & 38.8 & {\scriptsize$\pm$5.9} & 36.9 & {\scriptsize$\pm$5.2} & \textbf{36.8} & {\scriptsize$\pm$7.8} \\
Textures & 55.3 & 57.9 & {\scriptsize$\pm$7.0} & \underline{57.4} & {\scriptsize$\pm$3.7} & 58.9 & {\scriptsize$\pm$3.2} \\
360OpenSet & 57.7 & 43.0 & {\scriptsize$\pm$7.6} & 39.8 & {\scriptsize$\pm$6.3} & \textbf{39.6} & {\scriptsize$\pm$11.6} \\
Places & 58.7 & 54.7 & {\scriptsize$\pm$10.8} & \textbf{54.4} & {\scriptsize$\pm$9.6} & 66.9 & {\scriptsize$\pm$5.1} \\
COOD & 59.2 & 52.2 & {\scriptsize$\pm$6.0} & 53.2 & {\scriptsize$\pm$4.7} & \textbf{45.9} & {\scriptsize$\pm$9.1} \\
ImageNet-O & 60.8 & 56.1 & {\scriptsize$\pm$4.5} & \textbf{54.4} & {\scriptsize$\pm$3.6} & 55.6 & {\scriptsize$\pm$4.7} \\
SSB easy & 66.7 & 65.3 & {\scriptsize$\pm$3.8} & \textbf{63.3} & {\scriptsize$\pm$2.3} & 67.1 & {\scriptsize$\pm$3.7} \\
Species & 77.8 & 73.1 & {\scriptsize$\pm$9.1} & 79.9 & {\scriptsize$\pm$4.8} & \textbf{64.0} & {\scriptsize$\pm$14.7} \\
SSB hard & 78.2 & 74.7 & {\scriptsize$\pm$8.9} & 80.4 & {\scriptsize$\pm$4.4} & \textbf{67.1} & {\scriptsize$\pm$9.7} \\
IN1k-OOD & 82.0 & 82.4 & {\scriptsize$\pm$12.8} & 92.8 & {\scriptsize$\pm$5.9} & \textbf{67.0} & {\scriptsize$\pm$17.0} \\
\bottomrule
\end{tabular*}
\end{table}

\begin{table}[t]
\centering
\small
\setlength{\tabcolsep}{0pt}
\renewcommand{\arraystretch}{1.0}
\caption{
    \textbf{Negative vocabularies for \bench{}.}
    FPR@95 (\%, $\downarrow$) of NegLabel with the four vocabularies of \cref{tab:neg-sources}, as a mean over the eight VLMs, on the eleven popular OOD datasets at full size (\cref{tab:supp-full-vs-clean}), on NINCO and ImageNet-OOD, and on \bench{}.
    Best of the three mined vocabularies in \textbf{bold}.
    \emph{oracle} (gray) is a non-deployable empirical lower bound on FPR@95 and exists only for the datasets with labels available for every image.
}
\label{tab:supp-popular-vocab}
\begin{tabular*}{\columnwidth}{@{}l@{\extracolsep{\fill}}rrrrr@{}}
\toprule
Dataset & $n$ & WN & GPT & WN+GPT & oracle \\
\midrule
iNat-OOD Plants & 10\,000 & \textbf{1.2} & 40.4 & 5.2 & -- \\
OpenImage-O & 17\,632 & \textbf{29.6} & 55.9 & 35.3 & -- \\
360OpenSet & 11\,800 & 25.7 & 47.4 & \textbf{25.6} & \g{4.9} \\
COOD & 10\,350 & 36.0 & 50.0 & \textbf{33.7} & \g{9.5} \\
ImageNet-O & 2\,000 & 57.6 & 53.4 & \textbf{47.6} & \g{19.0} \\
Textures & 5\,640 & \textbf{66.4} & 72.4 & 67.9 & \g{7.5} \\
Places & 10\,000 & \textbf{47.7} & 73.4 & 63.1 & -- \\
SSB easy & 50\,000 & 67.8 & 66.0 & \textbf{61.8} & \g{37.4} \\
SSB hard & 49\,000 & 62.3 & 61.8 & \textbf{57.6} & \g{22.9} \\
Species & 10\,000 & \textbf{43.0} & 68.8 & 56.0 & -- \\
IN1k-OOD & 50\,000 & 65.4 & 63.1 & \textbf{60.9} & \g{21.7} \\
\midrule
NINCO & 5\,879 & 44.9 & 50.4 & \textbf{40.7} & \g{5.3} \\
ImageNet-OOD & 31\,802 & 66.3 & 55.5 & \textbf{53.6} & \g{17.0} \\
\midrule
\textit{mean} & & \textit{47.2} & \textit{58.3} & \textit{\textbf{46.8}} & \g{16.1} \\
\midrule
\textbf{\bench{}} & 655 & 74.6 & \textbf{58.6} & 58.8 & \g{13.8} \\
\bottomrule
\end{tabular*}
\end{table}

\section{Ablation Study: Size of the OOD Dataset}
\label{supp:size}

\begin{table*}[t]
\centering
\normalsize
\renewcommand{\arraystretch}{1.0}
\caption{
\textbf{Sub-sampling other OOD datasets down to the size of \bench{}.}
Each dataset is re-evaluated on 1000 random draws of \nbench{}~images without replacement, with the ID set and the detector fixed across draws (EVA02-L-448 with its per-dataset best of the 21 post-hoc detectors, named in gray).
\emph{full}: FPR@95 (\%, $\downarrow$) on all images;
\emph{mean}, \emph{bias}: average FPR@95 over the draws and its signed difference from \emph{full};
\emph{std}: standard deviation over the draws.
\emph{closed form}: the standard error per \cref{eq:fpr_variance} (square root of the variance) where $\widehat{p}_D := $ \textit{full} times the correction factor $\sqrt{(N{-}n)/(N{-}1)}$ that \cref{eq:fpr_variance} omits (for \bench{}, $\widehat{p}_D$ becomes 0.65 from the only pass).
$\rho$:~Spearman correlation between the detector ranking on a draw and on the full set.
Sub-sampling to \nbench{} images thus shifts neither FPR@95 nor the ranking, with sampling noise below 2 points.
The noise spans 1.39--1.80 points against a 12.7--16.4-point spread across datasets over the eight backbones of \cref{tab:supervised} evaluated in \cref{supp:size}.
}
\label{tab:supp-size}
\begin{tabular*}{\textwidth}{@{}l@{\extracolsep{\fill}} r S[table-format=3.1] r@{\extracolsep{2pt}\,}l@{\extracolsep{\fill}} r r r S[table-format=1.2] r@{}}
\toprule
\multirow[b]{2}{*}{Dataset} &
\multicolumn{1}{c}{\multirow[b]{2}{*}{$N$}} &
\multicolumn{1}{c}{\multirow[b]{2}{*}{$\nbench{}/N$ (\%)}} &
\multicolumn{2}{c}{\multirow[b]{2}{*}{full}} &
\multicolumn{5}{c}{1000 random draws of \nbench{}~samples} \\
\cmidrule(lr){6-10}
& & & & &
\multicolumn{1}{c}{mean} & \multicolumn{1}{c}{bias} & \multicolumn{1}{c}{std} &
\multicolumn{1}{c}{closed form} & \multicolumn{1}{c}{$\rho$} \\
\midrule
iNat-OOD Plants & 10\,000 &  6.6 &  1.21 & \g{ViM}     &  1.21 & \phantom{+}0.00 & 0.43 & 0.41 & 0.970 \\
OpenImage-O     & 17\,632 &  3.7 &  5.78 & \g{Maha++}  &  5.81 & +0.03           & 0.90 & 0.89 & 0.977 \\
360OpenSet      & 11\,800 &  5.6 & 11.95 & \g{E+ReAct} & 12.01 & +0.06           & 1.21 & 1.23 & 0.969 \\
COOD            & 10\,350 &  6.3 & 13.82 & \g{Maha++}  & 13.81 & \textminus0.01  & 1.30 & 1.31 & 0.952 \\
ImageNet-O      &  2\,000 & 32.8 & 18.00 & \g{Maha++}  & 18.06 & +0.06           & 1.23 & 1.23 & 0.974 \\
Textures        &  5\,640 & 11.6 & 21.18 & \g{Maha++}  & 21.12 & \textminus0.06  & 1.47 & 1.50 & 0.977 \\
Places          & 10\,000 &  6.6 & 25.27 & \g{NNGuide}     & 25.24 & \textminus0.03  & 1.68 & 1.64 & 0.984 \\
SSB easy        & 50\,000 &  1.3 & 30.04 & \g{NECO}    & 30.11 & +0.07           & 1.77 & 1.78 & 0.957 \\
SSB hard        & 49\,000 &  1.3 & 33.13 & \g{ViM}     & 33.18 & +0.05           & 1.82 & 1.83 & 0.978 \\
Species         & 10\,000 &  6.6 & 40.25 & \g{Maha}    & 40.26 & +0.01           & 1.85 & 1.85 & 0.978 \\
IN1k-OOD        & 50\,000 &  1.3 & 48.17 & \g{NAC}     & 48.16 & \textminus0.01  & 1.89 & 1.94 & 0.985 \\
NINCO           &  5\,879 & 11.1 & 16.89 & \g{Maha++}  & 16.90 & +0.01           & 1.36 & 1.38 & 0.980 \\
ImageNet-OOD    & 31\,802 &  2.1 & 18.80 & \g{NECO}    & 18.74 & \textminus0.06  & 1.52 & 1.51 & 0.957 \\
CLEAN-OOD       &  2\,715 & 24.1 & 12.15 & \g{Maha++}  & 12.19 & +0.04           & 1.09 & 1.11 & 0.955 \\
\midrule
\textbf{\bench{}} & \nbench{} & 100 & \textbf{65.0} & \g{Maha++} & -- & -- & -- & 1.86 & -- \\
\bottomrule
\end{tabular*}
\end{table*}

\Cref{sec:dataset_size} argues that, for a fixed model, detector, ID set, and ID-derived threshold, the OOD sample size affects the precision of FPR@95 but not its expected value.
The decision threshold is the 5th percentile of the \nid{} ID scores and is independent of the OOD set.
To test this empirically, we draw 1000 random \nbench{}-image subsets, matching the size of \bench{}, from each of the 14 external OOD datasets in \cref{tab:eva-siglip-subsets}.
We repeat the experiment for all eight supervised backbones.
For each combination of backbone and dataset, the detector with the lowest FPR@95 on the full dataset is selected once and then kept fixed across all draws.

\begin{figure}[t]
\centering
\includegraphics[width=.95\columnwidth]{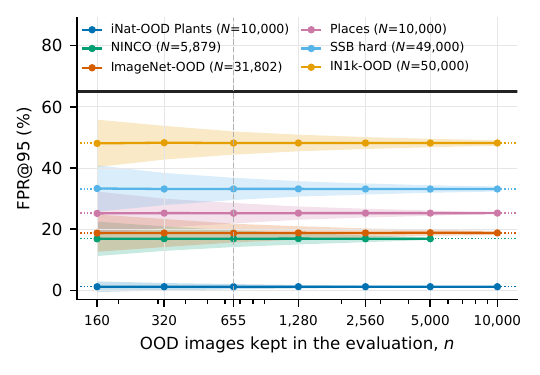}
\caption{\textbf{FPR@95 on random subsets and full OOD datasets.}
Six datasets from \cref{tab:eva-siglip-subsets}, each evaluated on 1000 random $n$-image draws using EVA02-L-448.
For each dataset, the best of the 21 post-hoc detectors is selected on the full dataset and then fixed across all draws; the ID set is also fixed.
The plot shows the mean FPR@95 (\%, $\downarrow$) over the draws, with bands indicating $\pm2$ standard deviations.
Dotted lines show the full-dataset values, the solid black line shows \bench{} as a fixed reference, and the gray vertical line marks $n{=}\nbench{}$.
The mean remains centered on the full-dataset value at every $n$.}
\label{fig:supp-size}
\end{figure}

\noindent\textbf{Unbiasedness (\cref{tab:supp-size,fig:supp-size}).}
For a fixed detector and ID-derived threshold, the sample FPR@95 is an unbiased estimator of the full-dataset rate.
At $n{=}\nbench{}$, the mean over the draws differs from the full-dataset rate by at most 0.13 points across all 112 combinations of backbone and dataset, even when a draw retains only 1.3\% of the dataset.

The draws are sampled without replacement from a finite dataset of size $N$.
Consequently, the standard deviation predicted using \cref{eq:fpr_variance} has to be multiplied by the finite-population correction
$$
    \sqrt{\frac{N-n}{N-1}}.
$$
Since this factor is always at most one, that value still remains a conservative upper bound.
The observed standard deviation agrees with the finite-population-corrected value to within 0.10 points.
Across all values of $n$ in \cref{fig:supp-size}, the mean remains centered on the full-dataset rate while the uncertainty band narrows approximately as $1/\sqrt{n}$.

The analytical standard deviation of 1.86 percentage points reported for \bench{} in \cref{tab:supp-size} justifies a distinct interpretation.
Because no larger finite pool of \bench{} images is available from which to perform sub-sampling, this quantity is instead a plug-in estimate derived from \cref{eq:fpr_variance}, computed with $n{=}\nbench{}$ and the observed FPR@95 of 65.0\%.
Consequently, it quantifies uncertainty under an assumption of independent sampling from the underlying OOD-generating distribution, rather than the variability induced by sub-sampling a fixed benchmark dataset.

\noindent\textbf{Resolution of a \nbench{}-image draw.}
For the ranking analysis, we evaluate all 21 detectors on every draw and compare their ranking with the ranking on the full dataset.
The mean Spearman correlation ranges from 0.95 to 1.00 for all combinations of backbone and dataset, and from 0.952 to 0.985 for EVA02-L-448 in \cref{tab:supp-size}.
A \nbench{}-image draw therefore largely preserves the detector ranking, although it may not reliably distinguish its top entries.
For \bench{}, the two best detectors differ by at most 2.6 points across the eight backbones, which is comparable to the sampling uncertainty.
\Cref{tab:eva-siglip-subsets,tab:supervised} therefore report the strongest result from the evaluated pool rather than a definitive ranking of its top detectors.

\noindent\textbf{Dataset effect vs.\ sampling noise.}
As summarized in \cref{sec:dataset_size}, the between-dataset spread ranges from 12.7 to 16.4 points across the eight backbones.
This is between 7.1 and 11.0 times the mean sampling standard deviation of a \nbench{}-image draw, which ranges from 1.39 to 1.80 points.
Thus, for the models and datasets evaluated, the choice of OOD dataset has a substantially larger effect than the sampling variation associated with using \nbench{} images.
In particular, the 16.8-point gap between \bench{} and the hardest external dataset (\cref{tab:supp-size}) is approximately nine times the 1.86-point plug-in standard deviation estimated in \cref{sec:dataset_size}.

\section{Additional Dataset Examples}
\label{supp:gallery}

Following on \cref{fig:teaser}, \cref{fig:supp-gallery} presents 42 additional examples from \bench{}.
For visual balance, half of the displayed examples are animals; overall, animals constitute 58.0\% of the complete dataset (based on intersection of the original \imnet[1] labels and the respective WordNet subtree~\cite{miller1995wordnet}).
Many certified labels are semantically close to the original class, such as \emph{tayra} versus \emph{weasel}, \emph{black-footed cat} versus \emph{lynx}, \emph{concert tuba} versus \emph{French horn}, and \emph{diesel locomotive} versus \emph{electric locomotive}; these examples illustrate the fine-grained failure mode.
The remaining examples illustrate a broader semantic mismatch, in which the depicted concept has no close \imnet[1] class and the original label reflects a scene-, material-, shape-, or contextual association: \emph{cloud-filled sky} labeled as \emph{wool}, \emph{small waterfall} as \emph{valley}, \emph{hexagonal pattern} as \emph{honeycomb}.

Because all images originate from the \imnet[1] collection pipeline, the benchmark is designed to minimize systematic differences in style, resolution, and capture conditions between ID and OOD data.
This shared origin does not eliminate all covariate differences, but it makes semantic category mismatch a more salient component of the evaluation.
\providecommand{\g}[1]{{\scriptsize\color{gray}#1}}
\definecolor{fprbar}{RGB}{176,203,231}
\begin{table*}[t]
\centering\footnotesize
\setlength{\tabcolsep}{4pt}
\renewcommand{\arraystretch}{1.0}
\caption{\textbf{Supervised methods - performance on \bench{}} ($n_{\text{OOD}}=\nbench{}$,
$n_{\text{ID}}=\nid{}$). Best FPR@95 ($\downarrow$) and AUROC ($\uparrow$) per model
of the same 21 post-hoc detectors as \cref{tab:supervised}; grey text names the
best-performing detector.
\emph{Pre-training} names the corpus or objective seen before the \imnet[1]
fine-tune (SSL rows are frozen backbones with a linear probe). The Top-1 is multi-label top-1 on the reannotated ID split,
both as in \cref{tab:supervised}. Bars are a reading aid drawn from a 64.4-point
baseline (shorter is better); they encode the same number reported on the right.
Gray shading: models from \cref{tab:supervised}. For space reasons, the main paper
reports only this representative subset of the models. $\dagger$~marks a
linear probe on a frozen backbone, as in \cref{tab:supervised}.
Checkpoints are in
\cref{tab:supp-checkpoints}.}
\label{tab:supp-other-models}
\begin{tabular}{@{}l l r r l@{\hspace{4pt}} r@{\hspace{4pt}} l
    @{\hspace{8pt}} r@{\hspace{4pt}} l@{}}
\toprule
Model & Pre-training & Params & Top-1 & & \multicolumn{2}{l}{FPR@95 $\downarrow$} &
\multicolumn{2}{l@{}}{AUROC $\uparrow$} \\
\midrule
EffNet-B0 & \imnet[1] & 5.3M & 81.6 & \makebox[18.0mm][r]{\textcolor{fprbar}{\rule{16.11mm}{3.2pt}}} & 82.9 & \g{Energy} & 72.5 & \g{rMaha++} \\
BiT-s & \imnet[1] & 44.5M & 81.8 & \makebox[18.0mm][r]{\textcolor{fprbar}{\rule{15.70mm}{3.2pt}}} & 82.4 & \g{rMaha} & 74.0 & \g{rMaha} \\
DeiT3-B-224 & \imnet[1] & 86.6M & 87.1 & \makebox[18.0mm][r]{\textcolor{fprbar}{\rule{14.19mm}{3.2pt}}} & 80.8 & \g{rMaha++} & 74.0 & \g{rMaha++} \\
XCiT-M-224 & \imnet[1] & 84.4M & 85.7 & \makebox[18.0mm][r]{\textcolor{fprbar}{\rule{13.91mm}{3.2pt}}} & 80.5 & \g{rMaha++} & 73.3 & \g{rMaha++} \\
\rowcolor{gray!30} ResNet50 & \imnet[1] & 25.6M & 84.9 & \makebox[18.0mm][r]{\textcolor{fprbar}{\rule{13.64mm}{3.2pt}}} & 80.2 & \g{rMaha++} & 74.2 & \g{rMaha++} \\
ViT-B-384 & \imnet[1] & 86.9M & 84.6 & \makebox[18.0mm][r]{\textcolor{fprbar}{\rule{13.23mm}{3.2pt}}} & 79.7 & \g{rMaha++} & 74.6 & \g{rMaha} \\
DeiT3-B-224-21k & \imnet[21] & 86.6M & 88.9 & \makebox[18.0mm][r]{\textcolor{fprbar}{\rule{13.09mm}{3.2pt}}} & 79.5 & \g{NAC} & 77.3 & \g{rMaha++} \\
DeiT3-B-384 & \imnet[1] & 86.9M & 88.1 & \makebox[18.0mm][r]{\textcolor{fprbar}{\rule{12.26mm}{3.2pt}}} & 78.6 & \g{rMaha++} & 75.5 & \g{rMaha++} \\
XCiT-M-224-d & \imnet[1] & 84.4M & 87.7 & \makebox[18.0mm][r]{\textcolor{fprbar}{\rule{11.85mm}{3.2pt}}} & 78.2 & \g{Maha++} & 74.8 & \g{rMaha++} \\
ViT-B-384-l2b & CLIP & 86.9M & 89.6 & \makebox[18.0mm][r]{\textcolor{fprbar}{\rule{10.89mm}{3.2pt}}} & 77.1 & \g{Maha++} & 78.9 & \g{rMaha++} \\
ViT-B-384-oai & CLIP & 86.9M & 89.1 & \makebox[18.0mm][r]{\textcolor{fprbar}{\rule{10.89mm}{3.2pt}}} & 77.1 & \g{rMaha++} & 78.4 & \g{rMaha++} \\
DeiT3-B-384-21k & \imnet[21] & 86.9M & 89.6 & \makebox[18.0mm][r]{\textcolor{fprbar}{\rule{10.89mm}{3.2pt}}} & 77.1 & \g{rMaha++} & 78.4 & \g{rMaha++} \\
SwinV2-B-256 & \imnet[1] & 87.9M & 87.6 & \makebox[18.0mm][r]{\textcolor{fprbar}{\rule{10.75mm}{3.2pt}}} & 76.9 & \g{rMaha++} & 74.1 & \g{rMaha++} \\
EffNetV2-M & \imnet[1] & 54.1M & 88.3 & \makebox[18.0mm][r]{\textcolor{fprbar}{\rule{10.75mm}{3.2pt}}} & 76.9 & \g{rMaha++} & 76.5 & \g{rMaha++} \\
ConvNeXt-B & \imnet[1] & 88.6M & 87.6 & \makebox[18.0mm][r]{\textcolor{fprbar}{\rule{9.93mm}{3.2pt}}} & 76.0 & \g{rMaha++} & 74.2 & \g{rMaha++} \\
BiT-m & \imnet[21] & 44.5M & 86.0 & \makebox[18.0mm][r]{\textcolor{fprbar}{\rule{9.38mm}{3.2pt}}} & 75.4 & \g{rMaha} & 78.2 & \g{rMaha++} \\
EffNet-b7-ns & JFT-300M (ns) & 66.3M & 89.6 & \makebox[18.0mm][r]{\textcolor{fprbar}{\rule{9.38mm}{3.2pt}}} & 75.4 & \g{rMaha++} & 77.2 & \g{rMaha++} \\
EffNetV2-M-21k & \imnet[21] & 54.1M & 89.2 & \makebox[18.0mm][r]{\textcolor{fprbar}{\rule{9.10mm}{3.2pt}}} & 75.1 & \g{rMaha++} & 79.6 & \g{rMaha++} \\
EffNet-b7 & \imnet[1] & 66.3M & 88.1 & \makebox[18.0mm][r]{\textcolor{fprbar}{\rule{8.55mm}{3.2pt}}} & 74.5 & \g{rMaha++} & 77.7 & \g{rMaha++} \\
ConvNeXt-T-21k & \imnet[21] & 28.6M & 87.6 & \makebox[18.0mm][r]{\textcolor{fprbar}{\rule{8.14mm}{3.2pt}}} & 74.0 & \g{NAC} & 77.6 & \g{Maha++} \\
ViT-B-384-oai-12k & CLIP${+}$\imnet[12] & 86.9M & 89.7 & \makebox[18.0mm][r]{\textcolor{fprbar}{\rule{8.00mm}{3.2pt}}} & 73.9 & \g{NAC} & 81.0 & \g{rMaha++} \\
ViT-B-384-l2b-12k & CLIP${+}$\imnet[12] & 86.9M & 89.8 & \makebox[18.0mm][r]{\textcolor{fprbar}{\rule{7.59mm}{3.2pt}}} & 73.4 & \g{rMaha++} & 81.1 & \g{rMaha++} \\
\rowcolor{gray!30} SwinV2-B-256-21k & \imnet[21] & 87.9M & 89.2 & \makebox[18.0mm][r]{\textcolor{fprbar}{\rule{7.45mm}{3.2pt}}} & 73.3 & \g{NAC} & 77.9 & \g{rMaha++} \\
ViT-B-224-21k & \imnet[21] & 86.6M & 87.7 & \makebox[18.0mm][r]{\textcolor{fprbar}{\rule{5.12mm}{3.2pt}}} & 70.7 & \g{rMaha++} & 80.1 & \g{Maha} \\
\rowcolor{gray!30} ConvNeXt-B-21k & \imnet[21] & 88.6M & 89.1 & \makebox[18.0mm][r]{\textcolor{fprbar}{\rule{3.74mm}{3.2pt}}} & 69.2 & \g{rMaha++} & 80.3 & \g{rMaha++} \\
\rowcolor{gray!30} ViT-B-384-21k & \imnet[21] & 86.9M & 89.3 & \makebox[18.0mm][r]{\textcolor{fprbar}{\rule{3.61mm}{3.2pt}}} & 69.0 & \g{Maha++} & 81.4 & \g{Maha} \\
\rowcolor{gray!30} DINOv2-g/14 $\dagger$ & SSL (LVD-142M) & 1.1B & 89.1 & \makebox[18.0mm][r]{\textcolor{fprbar}{\rule{3.33mm}{3.2pt}}} & 68.7 & \g{ViM} & 77.5 & \g{GEN} \\
\rowcolor{gray!30} ConvNeXtV2-L & FCMAE${+}$\imnet[21] & 198.0M & 90.3 & \makebox[18.0mm][r]{\textcolor{fprbar}{\rule{2.52mm}{3.2pt}}} & 67.8 & \g{rMaha++} & 83.1 & \g{Maha++} \\
\rowcolor{gray!30} EVA02-L-448 & MIM${+}$\imnet[21] & 305.1M & \textbf{91.5} & \makebox[18.0mm][r]{\textcolor{fprbar}{\rule{0.51mm}{3.2pt}}} & 65.0 & \g{Maha++} & \textbf{84.2} & \g{rMaha++} \\
\rowcolor{gray!30} DINOv3-7B $\dagger$ & SSL (LVD-1689M) & 6.7B & 90.1 & \makebox[18.0mm][r]{\textcolor{fprbar}{\rule{0.00mm}{3.2pt}}} & \textbf{64.4} & \g{rMaha++} & 81.0 & \g{MCM} \\
\bottomrule
\end{tabular}
\end{table*}

\begin{table*}[t]
\centering\scriptsize
\setlength{\tabcolsep}{3pt}
\caption{\textbf{Checkpoints of all evaluated models.}
Top: the 30 supervised classifiers of \cref{tab:supp-other-models}, which include the eight of \cref{tab:supervised}, as \texttt{timm} checkpoints with explicit tags; $\dagger$ marks a frozen backbone with a linear probe trained on the \imnet[1] training split.
Bottom: the eight zero-shot VLMs of \cref{tab:vlm-posthoc} as \texttt{open\_clip} architecture\,/\,pre-training pairs with their input resolution, except AIMv2-L (Hugging Face).}
\label{tab:supp-checkpoints}
\label{tab:supp-main-checkpoints}
\begin{tabular}{@{}l l @{\hspace{8pt}} l l@{}}
\toprule
\multicolumn{4}{@{}l}{\textit{Supervised classifiers}} \\
Model & \texttt{timm} checkpoint & Model & \texttt{timm} checkpoint \\
\midrule
ViT-B-384-l2b-12k & \texttt{\scriptsize vit\_base\_patch16\_clip\_384.laion2b\_ft\_in12k\_in1k} & EffNet-b7-ns & \texttt{\scriptsize tf\_efficientnet\_b7.ns\_jft\_in1k} \\
ViT-B-384-oai-12k & \texttt{\scriptsize vit\_base\_patch16\_clip\_384.openai\_ft\_in12k\_in1k} & ViT-B-384 & \texttt{\scriptsize vit\_base\_patch16\_384.augreg\_in1k} \\
ViT-B-384-l2b & \texttt{\scriptsize vit\_base\_patch16\_clip\_384.laion2b\_ft\_in1k} & SwinV2-B-256 & \texttt{\scriptsize swinv2\_base\_window16\_256.ms\_in1k} \\
ViT-B-384-oai & \texttt{\scriptsize vit\_base\_patch16\_clip\_384.openai\_ft\_in1k} & DeiT3-B-384 & \texttt{\scriptsize deit3\_base\_patch16\_384.fb\_in1k} \\
ViT-B-384-21k & \texttt{\scriptsize vit\_base\_patch16\_384.augreg\_in21k\_ft\_in1k} & DeiT3-B-224 & \texttt{\scriptsize deit3\_base\_patch16\_224.fb\_in1k} \\
ViT-B-224-21k & \texttt{\scriptsize vit\_base\_patch16\_224.augreg\_in21k\_ft\_in1k} & XCiT-M-224 & \texttt{\scriptsize xcit\_medium\_24\_p16\_224.fb\_in1k} \\
SwinV2-B-256-21k & \texttt{\scriptsize swinv2\_base\_window12to16\_192to256.ms\_in22k\_ft\_in1k} & XCiT-M-224-d & \texttt{\scriptsize xcit\_medium\_24\_p16\_224.fb\_dist\_in1k} \\
DeiT3-B-384-21k & \texttt{\scriptsize deit3\_base\_patch16\_384.fb\_in22k\_ft\_in1k} & ConvNeXt-B & \texttt{\scriptsize convnext\_base.fb\_in1k} \\
DeiT3-B-224-21k & \texttt{\scriptsize deit3\_base\_patch16\_224.fb\_in22k\_ft\_in1k} & BiT-s & \texttt{\scriptsize resnetv2\_101x1\_bit} \\
ConvNeXt-B-21k & \texttt{\scriptsize convnext\_base.fb\_in22k\_ft\_in1k} & EffNetV2-M & \texttt{\scriptsize tf\_efficientnetv2\_m.in1k} \\
ConvNeXt-T-21k & \texttt{\scriptsize convnext\_tiny.fb\_in22k\_ft\_in1k\_384} & EffNet-b7 & \texttt{\scriptsize tf\_efficientnet\_b7.ra\_in1k} \\
BiT-m & \texttt{\scriptsize resnetv2\_101x1\_bit.goog\_in21k\_ft\_in1k} & EffNet-B0 & \texttt{\scriptsize efficientnet\_b0.ra\_in1k} \\
EffNetV2-M-21k & \texttt{\scriptsize tf\_efficientnetv2\_m.in21k\_ft\_in1k} & ResNet50 & \texttt{\scriptsize resnet50.a1\_in1k} \\
ConvNeXtV2-L & \texttt{\scriptsize convnextv2\_large.fcmae\_ft\_in22k\_in1k\_384} & DINOv2-g/14 $\dagger$ & \texttt{\scriptsize dinov2\_vitg14} + linear probe \\
EVA02-L-448 & \texttt{\scriptsize eva02\_large\_patch14\_448.mim\_m38m\_ft\_in22k\_in1k} & DINOv3-7B $\dagger$ & \texttt{\scriptsize dinov3\_vit7b16} + linear probe \\
\midrule
\multicolumn{4}{@{}l}{\textit{Zero-shot vision--language models}} \\
Model & \texttt{open\_clip} checkpoint, resolution & Model & \texttt{open\_clip} checkpoint, resolution \\
\midrule
MetaCLIP2-H/14 & \texttt{\scriptsize ViT-H-14-worldwide-378 / metaclip2\_worldwide}, 378\,px & CLIP-B/16 & \texttt{\scriptsize ViT-B-16-quickgelu / openai}, 224\,px \\
OpenCLIP-H/14 & \texttt{\scriptsize ViT-H-14-378-quickgelu / dfn5b}, 378\,px & CLIP-L/14 & \texttt{\scriptsize ViT-L-14-336-quickgelu / openai}, 336\,px \\
SigLIP-SO400M & \texttt{\scriptsize ViT-SO400M-14-SigLIP-384 / webli}, 384\,px & AIMv2-L & \texttt{\scriptsize apple/aimv2-large-patch14-224-lit}, 224\,px \\
SigLIP2-g & \texttt{\scriptsize ViT-gopt-16-SigLIP2-384 / webli}, 384\,px & PE-Core-L/14 & \texttt{\scriptsize PE-Core-L-14-336 / meta}, 336\,px \\
\bottomrule
\end{tabular}
\end{table*}

\section{Additional Supervised Models}
\label{supp:model-zoo}

\Cref{tab:supp-other-models} expands the supervised evaluation from the eight representative models in \cref{tab:supervised} to 30 classifiers, covering parameter counts from 5.3M to 6.7B, eight architecture families, and five pre-training regimes.
For each classifier, we report the best FPR@95 over the same 21 post-hoc detectors used in the main paper.
The lowest value is 64.4\%, obtained by the DINOv3-7B linear probe with relative Mahalanobis++.
Thus, even the strongest of the evaluated model and detector combinations accepts almost two thirds of the \bench{} images as ID at the threshold that accepts 95\% of the ID set.
None of the 22 additional classifiers improves on this minimum.
The expanded evaluation therefore shows that the high FPR@95 on \bench{} is not specific to the representative architectures selected for the main paper.

Pre-training nevertheless has a measurable effect within the evaluated model families.
Out of the 26 models, the mean best FPR@95 decreases from 79.0\% for \imnet[1]-only training to 75.4\% for JFT noisy-student pre-training and 73.7\% for both \imnet[21] pre-training and CLIP checkpoints with an intermediate \imnet[12] stage.
All seven matched model pairs favor \imnet[21] over \imnet[1] pre-training, with an average reduction of 4.7 percentage points.
Reannotated Top-1 is correlated with FPR@95 ($\rho{=}{-}0.61$), but does not fully determine it.
Adding an \imnet[12] stage between CLIP pre-training and \imnet[1] fine-tuning improves FPR@95 by 3.4 points while changing Top-1 by only 0.1 to 0.6 points.

\section{Implementation Details}
\label{supp:impl}

\noindent\textbf{Checkpoints.}
\Cref{tab:supp-checkpoints} lists every evaluated model, supervised and zero-shot, with its identifying checkpoint.
The DINOv2-g/14 and DINOv3-7B probes are single linear layers on the frozen 1536- and 4096-dimensional final embeddings, trained with cross-entropy on the \imnet[1] training split.

\noindent\textbf{Zero-shot scoring.}
Every label, ID or negative, is encoded with the eight templates of \cref{supp:prompts} and the $\ell_2$-normalized template embeddings are averaged; image and text embeddings are compared by cosine similarity, scaled by $1/\tau$ with $\tau{=}0.01$.
MCM is the maximum softmax over the ID names; MaxLogit, Energy, GEN, Entropy and Margin are computed from the same scaled similarities.
For EOE, we use the penalty weight of $\beta{=}0.25$ as in the release; sweeping $\beta\in\{0,0.25,0.5,1\}$ moves FPR@95 on \bench{} by at most 1--2 points.
GL-MCM~\cite{miyai2025glmcm} amends MCM with the maximum softmax over the patch-token logits, using weight $\lambda{=}1$, of a second forward pass in which the last block's attention output is replaced by its value branch.

\clearpage
\raggedbottom
\section{Prompts}
\label{supp:prompts}

\noindent\textbf{Zero-shot classification prompt templates.}
Every label is encoded with the eight prompt templates
\emph{``\{\}''},
\emph{``a photo of a \{\}.''},
\emph{``a bad photo of a \{\}.''},
\emph{``a photo of many \{\}.''},
\emph{``a photo of the large \{\}.''},
\emph{``a photo of the small \{\}.''},
\emph{``a cropped photo of a \{\}.''} and
\emph{``a close-up photo of a \{\}.''};
that is, the bare name and seven of the OpenAI ImageNet templates~\cite{radford2021clip}, the ensemble adopted also by~\cite{volkov2025vlmembeddings}.

\penalty-10000

\noindent\textbf{EOE OOD method.}
We use the three prompts released with EOE~\cite{cao2024eoe}.
In each, \texttt{[ID names]} stands for the semicolon-separated list of the 1000 ID class names.

\begin{promptbox}
\emph{Far-OOD:} ``I have gathered images of 1000 distinct categories: \texttt{[ID names]}. Summarize what broad categories these categories might fall into based on visual features. Now, I am looking to identify 500 classes that visually resemble these broad categories but have no direct relation to these broad categories. Please list these 500 categories for me.''
\end{promptbox}

\begin{promptbox}
\emph{Near-OOD:} ``Given the image categories: \texttt{[ID names]}. Please suggest visually similar categories that are not directly related or belong to the same primary group as these categories. Provide suggestions that share visual characteristics but are from broader and different domains than these categories. List 500 such categories.''
\end{promptbox}

\begin{promptbox}
\emph{Fine-grained:} ``I have a dataset containing 1000 different species of object. I need a list of 500 distinct object species that are NOT present in my dataset, and ensure there are no repetitions in the list you provide. For context, the species in my dataset are: \texttt{[ID names]}.''
\end{promptbox}

\begin{promptbox}
\emph{Output format (all three):} ``Output ONLY a single JSON object with one key \texttt{"categories"} mapping to a list of exactly 500 strings. Each string is a concrete, photographable category name, lowercase, at most 3 words. No duplicates. Add nothing else.''
\end{promptbox}

\penalty-10000

\noindent\textbf{GPT-5.4 prompt for NegLabel used to obtain ID siblings in \cref{tab:neg-labels}}.
The mining over the 1000 distinct ID class names (14 names are shared by classes merged in ReImageNet \cite{reimagenet}) is repeated ten times using the following prompt:
\begin{promptbox}
``You are building a hard-negative label vocabulary for open-set image recognition.
The in-distribution (ID) label space is ImageNet-1k. For EACH ID class name I give
you, produce 20 real-world category names that are EASILY VISUALLY CONFUSED with it
in a photograph -- similar shape, texture, colour, or typical setting -- but that are
NOT the same category and are NOT themselves one of the 1000 ImageNet-1k classes.

The goal is the `just outside the class boundary' shell: categories a strong image
classifier could plausibly mistake for the ID class, yet that genuinely lie outside
it. These become negative labels, so precision at the boundary matters more than
diversity.
\end{promptbox}
\begin{promptbox}
Rules:
(1) Concrete, photographable nouns, at most 3 words each, lowercase.
(2) For living things, give SIBLING species/breeds at the SAME taxonomic granularity
(e.g.\ ``german shepherd'' $\rightarrow$ ``dutch shepherd'', ``bohemian shepherd''; NOT the parent
``dog'', NOT a coarser or finer level, NOT the class itself).
(3) For man-made objects, give visually adjacent object types (e.g.\ ``cash machine''
$\rightarrow$ ``ticket kiosk'', ``self-service terminal'').
(4) No synonyms, plurals, or spelling variants of the target class.
(5) No duplicates within a class's list.
(6) Do NOT output any of the 20 neighbours as an existing ImageNet-1k class name.
(7) Aim for the near boundary, never far-away/unrelated objects.

Output ONLY a single JSON object mapping each input class name (verbatim, exactly as
given, including any parenthetical disambiguation) to its list of 20 strings.
Return every class in the input; add nothing else.''
\end{promptbox}
Labels that match an \imnet[1] class name or one of its WordNet synonyms are removed, resulting in 19{,}597--19{,}632 labels per generation. \\

\penalty-10000

\noindent\textbf{Annotation evidence for GPT-5.4.}
The open-vocabulary description shown to the annotators -- evidence source (A) of \cref{fig:pipeline}.
It is produced with GPT-5.4 through the OpenAI API, adapted from the MLLM pipeline of~\cite{kisel2026multimodal}. 
The model must either commit to one of the 1000 \imnet[1] class names (closed world) or state that the subject is not in the list and give a free-form fine-grained label (open world).
\texttt{[ID names]} stands for the semicolon-separated list of the 1000 \imnet[1] class names.
\begin{promptbox}
``You are an expert image classifier. You are provided with a list of allowed class names separated by semicolons (;).

Classification Rules:

Determine the Main Subject: Focus only on the single most dominant and relevant object in the image, even if multiple subjects are present.

List Matching: Evaluate if the main subject belongs to one of the provided class names.

Handling `Not in List' (Fine-Grained Alternative): If you are not sure or the subject does not belong to the provided list, you must state ``nothing from provided classnames'' and provide your top-1 alternative.

Alternative Rule 1: This alternative must be the most fine-grained, specific natural-language label possible (e.g., ``golden retriever puppy'', ``1950s red convertible'', ``blue morpho butterfly'', ``ceramic coffee mug with floral pattern''). Avoid generic terms like ``dog'', ``car'', or ``bird''. Reflect detailed visual distinctions like species, make/model, style, color, or material.

Alternative Rule 2: If the dominant object cannot be clearly identified, return a concise descriptive label of its appearance (e.g., ``abstract metal sculpture'', ``blurry human silhouette'').
\end{promptbox}
\begin{promptbox}
Output Rules:

Return exactly one output per image.

Do not include explanations, reasoning, conversational filler, or any extra text whatsoever.

Strictly format your output matching ONE of the following two templates:

Format A (Match from list): ClassName

Format B (No match from list): nothing from provided classnames, [FineGrainedAlternative]

Allowed class names (semicolon-separated): \texttt{[ID names]}''
\end{promptbox}

\newcommand{\suppcell}[3]{%
\begin{minipage}[t]{0.13771\linewidth}\centering
  \includegraphics[width=\linewidth]{figs/supp_gallery/#1.jpg}\\[-2pt]
  {\scriptsize\textcolor{green!50!black}{#2}}\\[-3.5pt]
  {\scriptsize\textcolor{red}{#3}}
\end{minipage}%
}

\begin{figure*}[p]\centering
  \begin{minipage}{\textwidth}\centering
  \suppcell{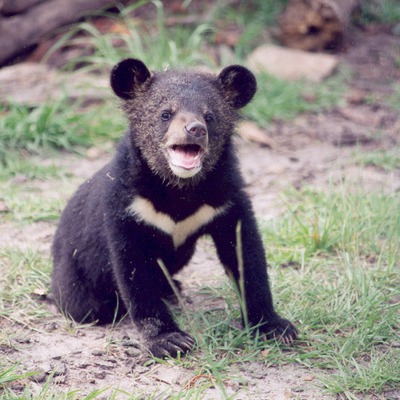}{Asiatic black bear cub}{American black bear}%
  \suppcell{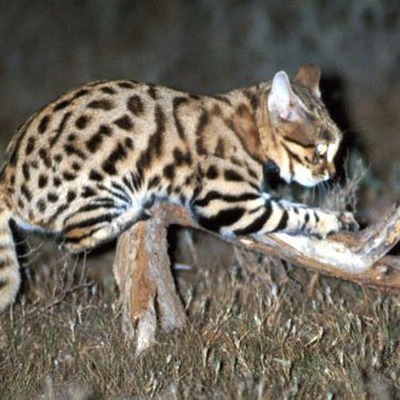}{black-footed cat}{lynx}%
  \suppcell{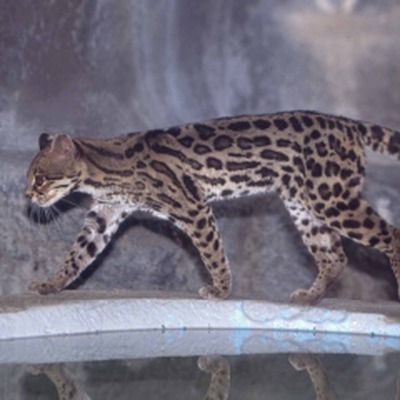}{ocelot}{jaguar}%
  \suppcell{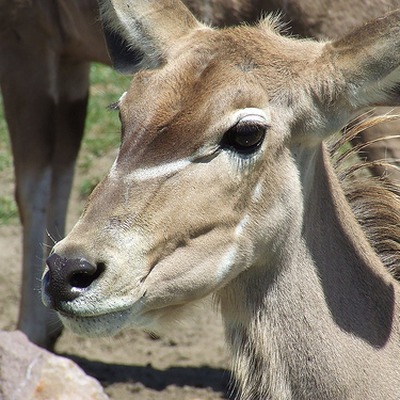}{kudu antelope}{gazelle}%
  \suppcell{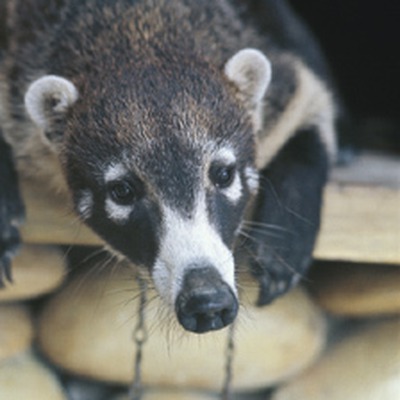}{white-nosed coati}{badger}%
  \suppcell{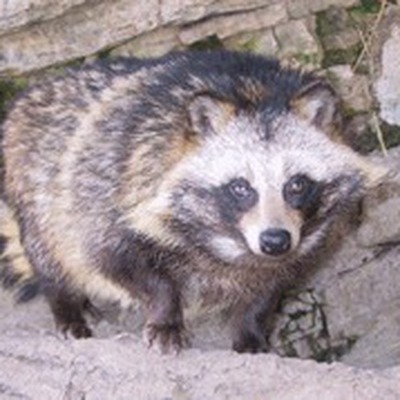}{raccoon dog}{coyote}%
  \suppcell{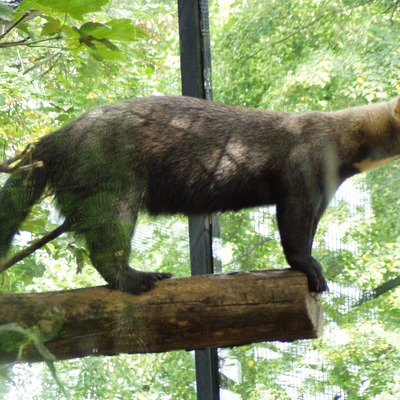}{tayra}{weasel}%
  \\[0.9ex]
  \suppcell{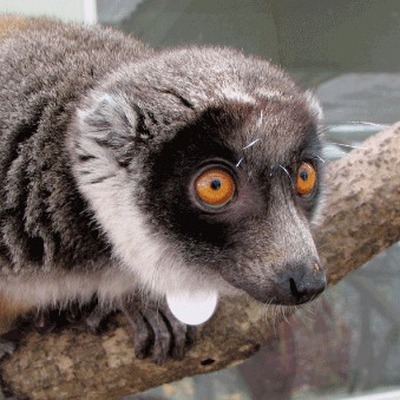}{mongoose lemur}{mongoose}%
  \suppcell{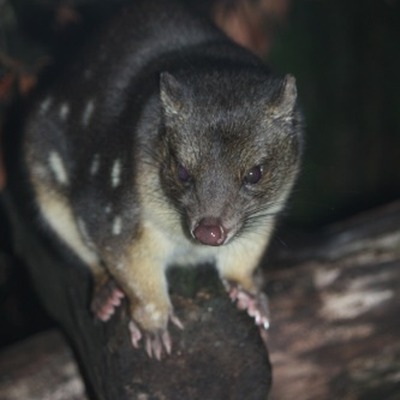}{quoll}{wombat}%
  \suppcell{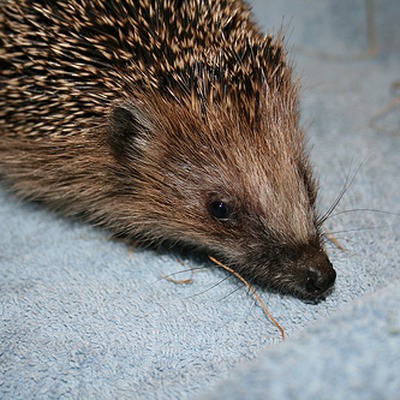}{hedgehog}{porcupine}%
  \suppcell{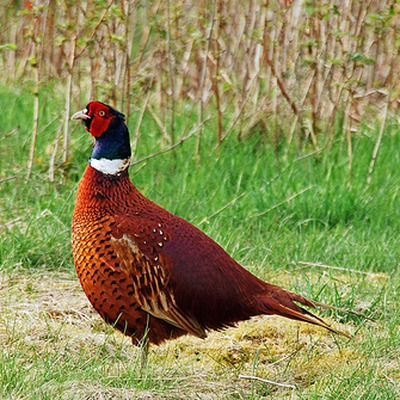}{ring-necked pheasant}{black grouse}%
  \suppcell{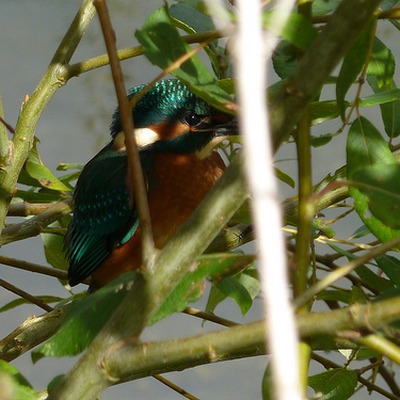}{common kingfisher}{American dipper}%
  \suppcell{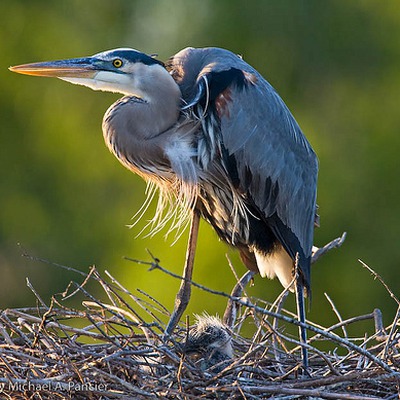}{great blue heron}{little blue heron}%
  \suppcell{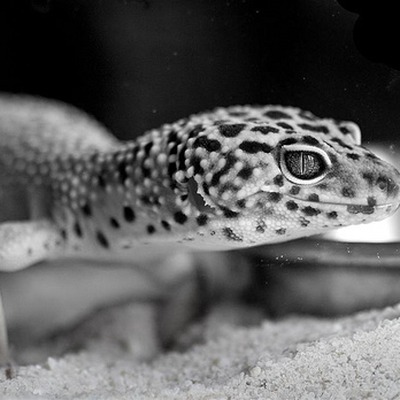}{leopard gecko}{banded gecko}%
  \\[0.9ex]
  \suppcell{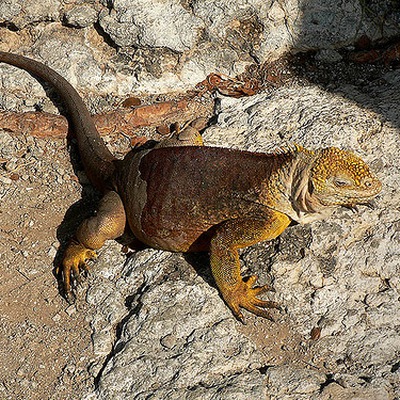}{Galapagos land iguana}{green iguana}%
  \suppcell{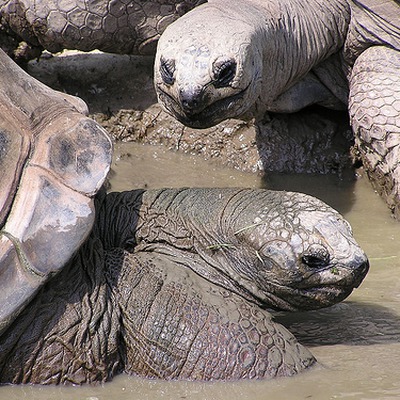}{giant tortoise}{mud turtle}%
  \suppcell{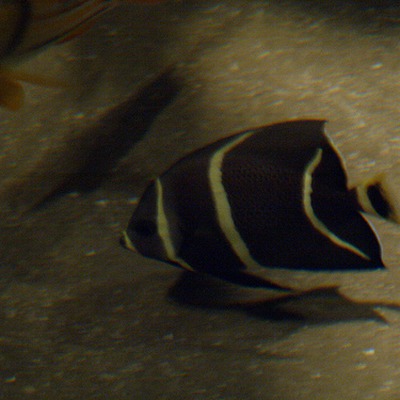}{French angelfish}{clownfish}%
  \suppcell{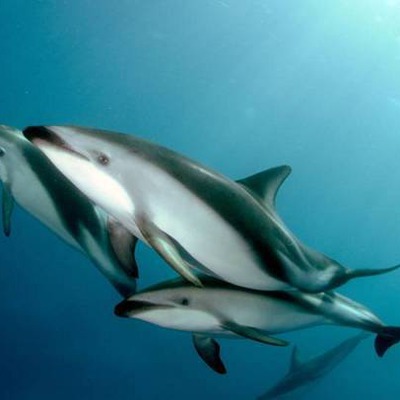}{dolphin}{snoek fish}%
  \suppcell{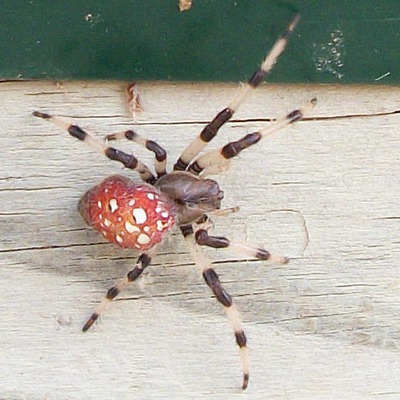}{four-spot orb weaver}{barn spider}%
  \suppcell{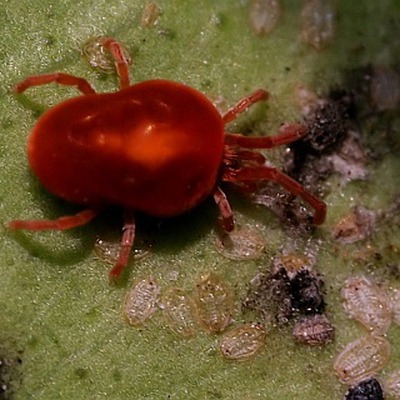}{red velvet mite}{tick}%
  \suppcell{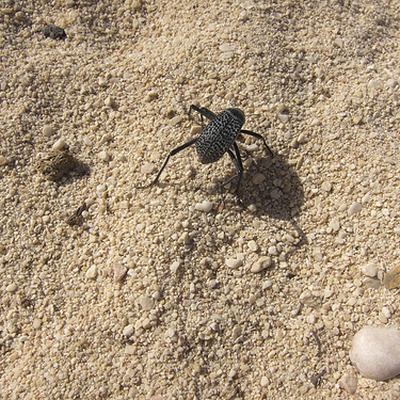}{darkling beetle}{dung beetle}%
  \\[0.9ex]
  \suppcell{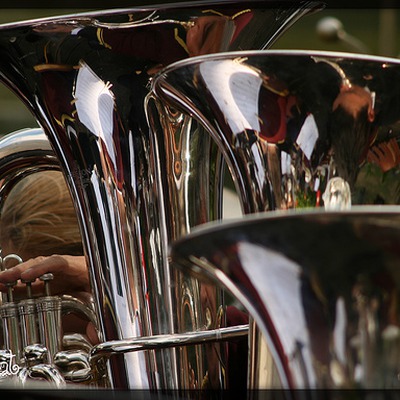}{concert tuba}{French horn}%
  \suppcell{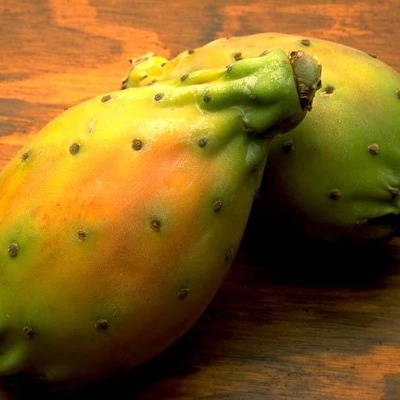}{pear cactus fruits}{fig}%
  \suppcell{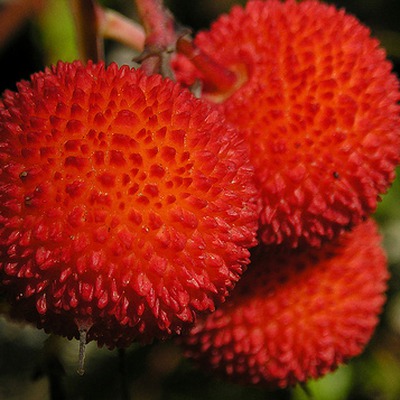}{strawberry tree fruit}{strawberry}%
  \suppcell{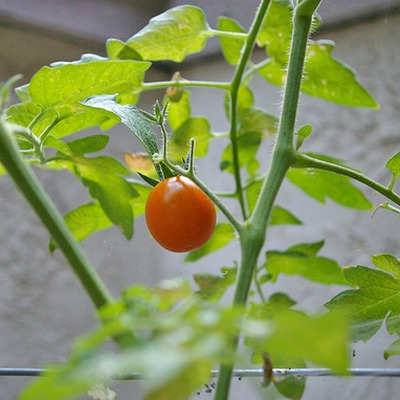}{cherry tomato}{orange}%
  \suppcell{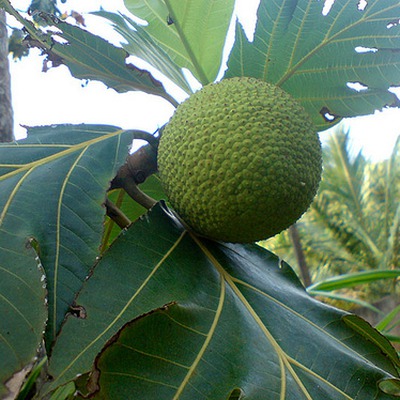}{breadfruit tree}{jackfruit}%
  \suppcell{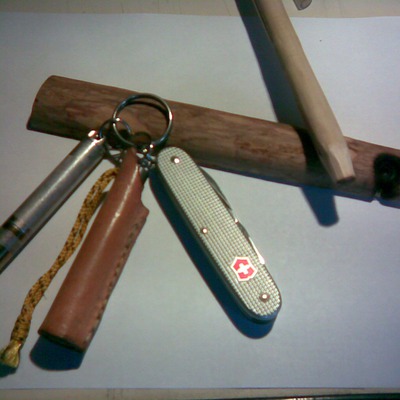}{swiss knife}{drumstick}%
  \suppcell{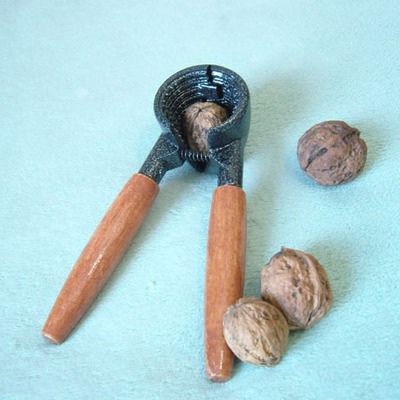}{walnut nutcracker}{can opener}%
  \\[0.9ex]
  \suppcell{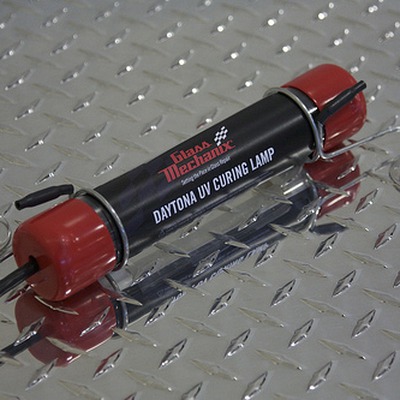}{UV curing lamp}{tool kit}%
  \suppcell{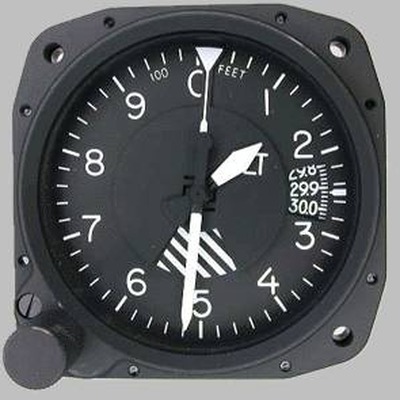}{altimeter}{barometer}%
  \suppcell{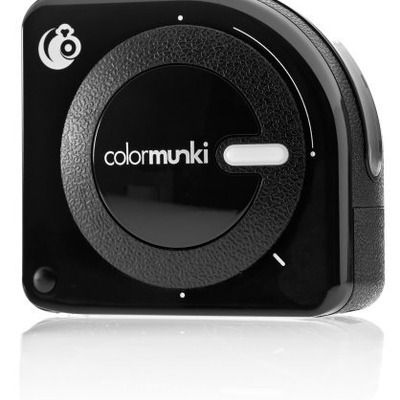}{spectrophotometer}{projector}%
  \suppcell{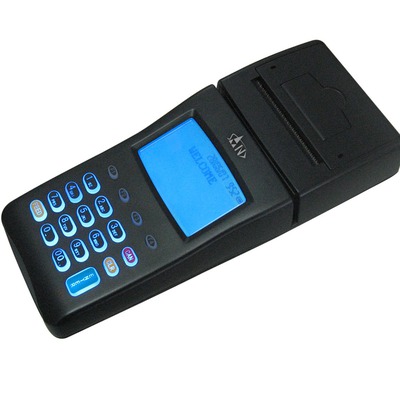}{payment terminal}{hand-held computer}%
  \suppcell{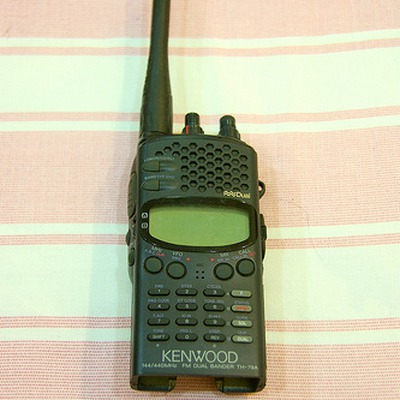}{walkie-talkie}{radio}%
  \suppcell{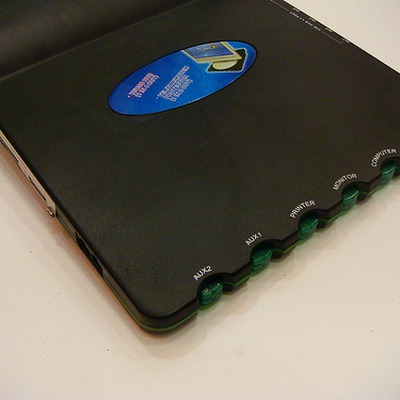}{overvoltage protector}{modem}%
  \suppcell{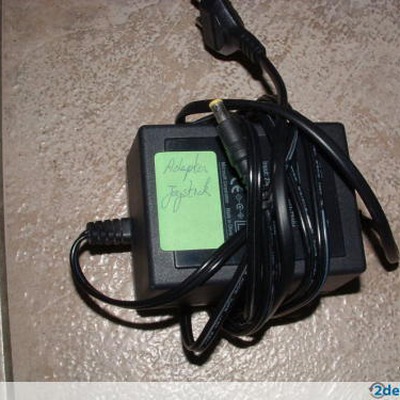}{power adapter}{joystick}%
  \\[0.9ex]
  \suppcell{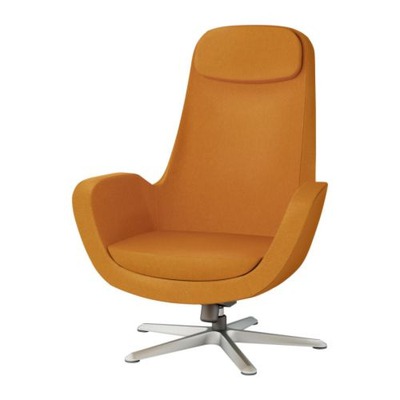}{lounge chair}{rocking chair}%
  \suppcell{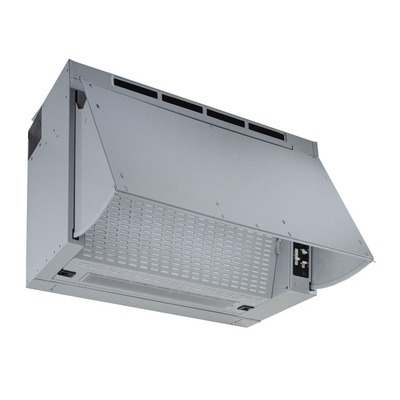}{range hood}{dishwasher}%
  \suppcell{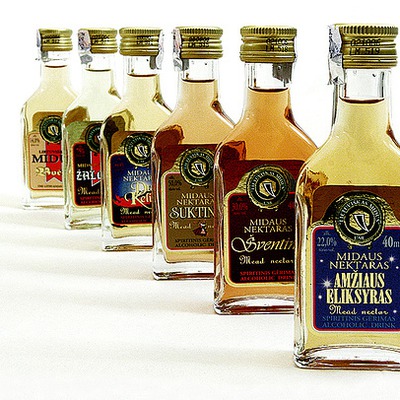}{mead bottle}{wine bottle}%
  \suppcell{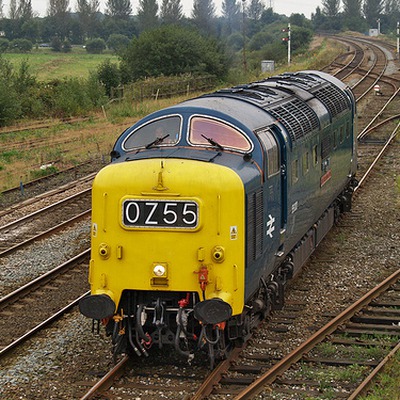}{diesel locomotive}{electric locomotive}%
  \suppcell{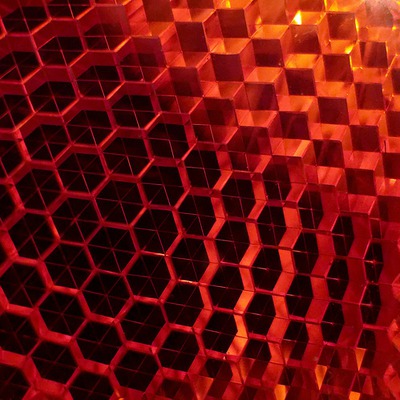}{hexagonal pattern}{honeycomb}%
  \suppcell{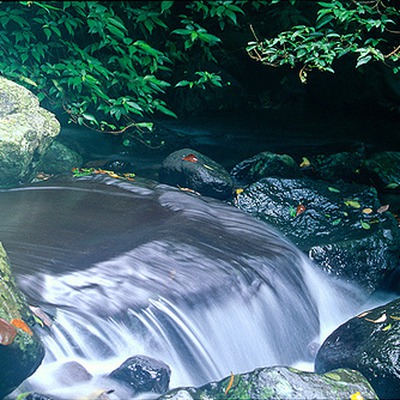}{small waterfall}{valley}%
  \suppcell{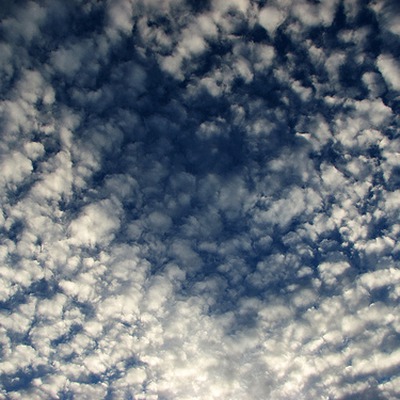}{cloud-filled sky}{wool}%
  \end{minipage}
  \caption{\textbf{Additional \bench{} examples.}
    42 OOD images with the \textcolor{green!50!black}{certified OOD label}, none of which is in the \imnet[1] class space, and the \textcolor{red}{original \imnet[1] label} the image was collected under.
    The top three rows show animals, the bottom three the remaining classes.
  }
  \label{fig:supp-gallery}
\end{figure*}

\end{document}